\documentclass[10pt,reqno]{amsart}

\usepackage{companypaper}

\usepackage{algorithm}
\usepackage{algorithmic}

\usepackage{natbib}

\usepackage{hyphenat}
\usepackage{enumitem}
\usepackage{verbatim}

\usepackage[T1]{fontenc}
\usepackage[utf8]{inputenc}

\usepackage{microtype}
\usepackage{inconsolata}
\usepackage{graphicx}
\usepackage{float}

\usepackage{amssymb}
\usepackage{mathtools}

\usepackage{booktabs}
\usepackage{colortbl}
\usepackage{multirow}
\usepackage{makecell}

\usepackage[most]{tcolorbox}

\usepackage{pmboxdraw}
\usepackage{xcolor}

\definecolor{nesytxt}{RGB}{20,83,45}    % green-900: NeSy row text only

\makeatletter
\pretocmd{\@settitle}{\let\uppercasenonmath\@gobble}{}{}
\patchcmd{\@settitle}{\bfseries}{\bfseries\LARGE}{}{}
\pretocmd{\@setauthors}{\let\MakeUppercase\@firstofone}{}{}
\patchcmd{\@setauthors}{\centering\footnotesize}{\centering\large}{}{}
\renewcommand{\@setaddresses}{}
\renewcommand{\@setdate}{%
  \noindent\normalfont\footnotesize
  \textsuperscript{1}Fujitsu Research India Private Limited,~
  Bengaluru, India\par
  \vskip 2pt
  \textsuperscript{*}Equal contribution.\par
  \vskip 2pt
  \textbf{Correspondence}: \textcolor{blue}{\texttt{\{sameer.malik, ayush.singh, amar.azad\}@fujitsu.com}}\par
}
\makeatother

\CompanyLogo{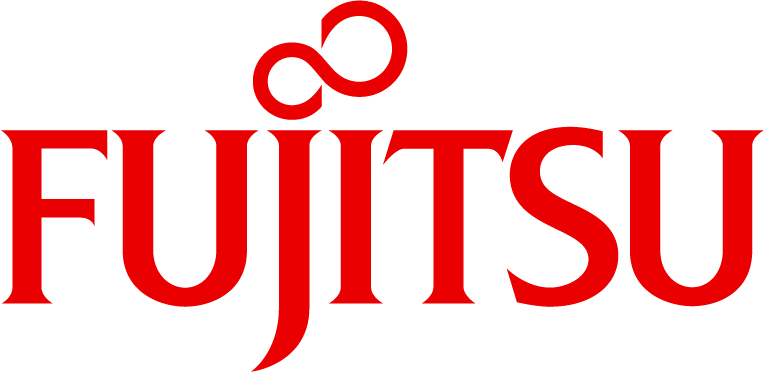}
\CompanyLogoHeight{11mm}
\CompanyHeaderText{Fujitsu Research India}
\CompanyDocType{}
\CompanyClassification{}
\CompanySubtitle{}
\CompanyRunningTitle{PolicyGuard: Neuro-Symbolic Compliance Review}

\title{PolicyGuard: From Organizational Policies to Neuro-Symbolic Compliance Review Engines}

\author{
Sameer Malik\textsuperscript{1,*},
Ayush Singh\textsuperscript{1,*},
Amar Prakash Azad\textsuperscript{1}}

\date{~}  % date field repurposed for affiliations via \@setdate above

\begin{document}

\maketitle

\begin{abstract}
Policy-grounded document review requires determining whether a target document complies with organization-specific policies, guidelines, or playbooks. While large language models can assist with policy interpretation and document analysis, end-to-end prompting leaves the applied policy logic implicit, making compliance decisions difficult to inspect, update, and test. We present \textsc{PolicyGuard}, a neuro-symbolic framework for policy-grounded document compliance review. PolicyGuard converts organizational policy guidance into an executable review engine consisting of typed relational logic rules and atom-level extraction questions. During review, LLMs answer these local questions using retrieved document evidence, and a symbolic evaluator applies the formal rules to detect non-compliance. We instantiate and evaluate PolicyGuard on company-specific NDA compliance review, where contract clauses must be checked against organization-specific negotiation policies. By separating policy formalization, local document interpretation, and symbolic compliance evaluation, PolicyGuard makes document review more explicit, maintainable, and systematically testable.

\end{abstract}

\begin{figure}[t]
    \centering
    \includegraphics[width=0.6\columnwidth]{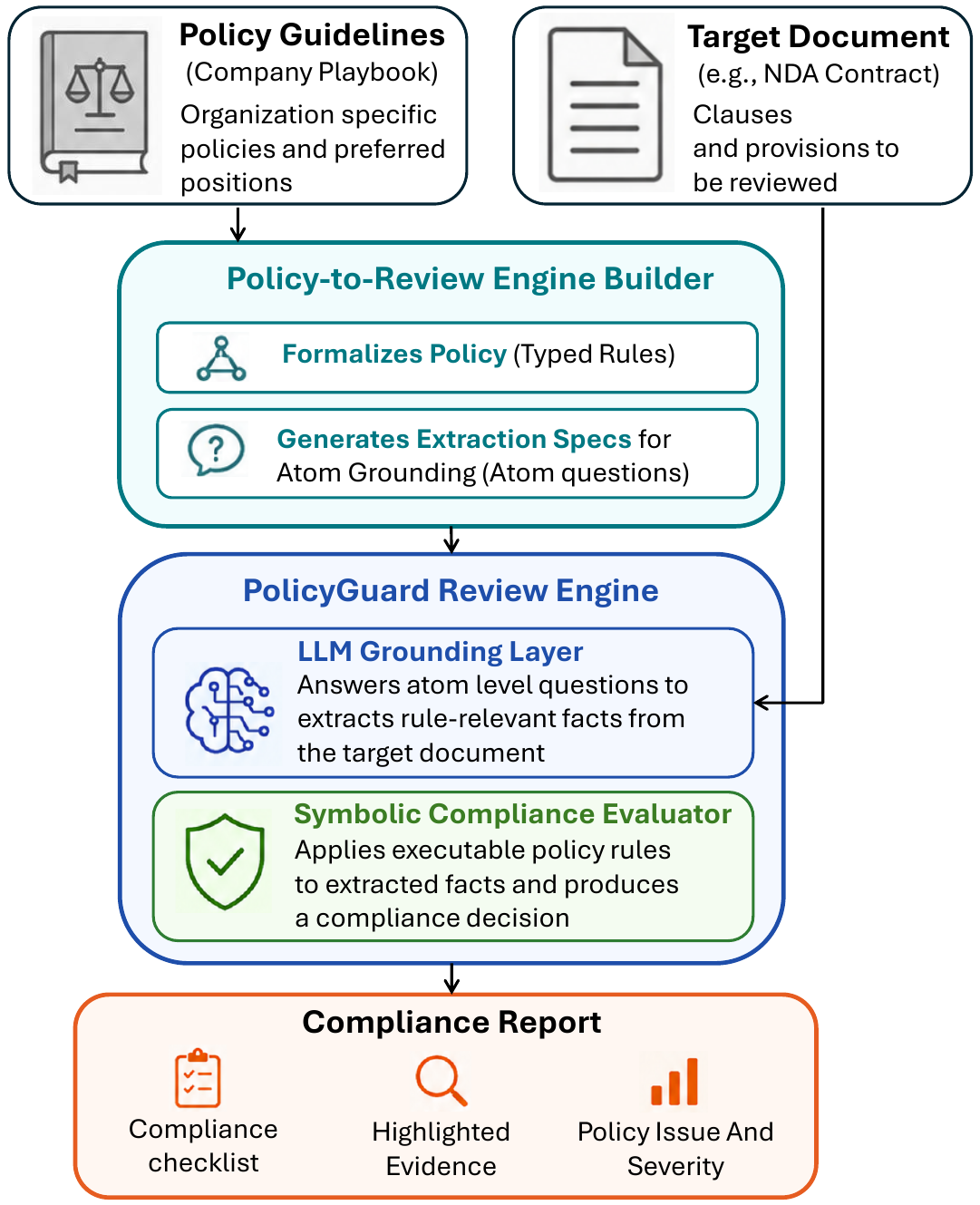}
    \caption{Overview of the \textsc{PolicyGuard}. Given organization-specific policy guidance and a target document, PolicyGuard constructs a review engine that formalizes policy conditions as typed rules and atom-level extraction questions. At review time, LLMs ground the required atoms in document evidence, while a symbolic evaluator applies the executable rules to produce an auditable compliance report.}
    \label{fig:overview}
\end{figure}

\section{Introduction}
\label{sec:introduction}

Organizations routinely need to determine whether documents comply with internal policies, guidelines, or playbooks. These policies may govern contracts, procurement documents, security procedures, operational reports, technical specifications, or other enterprise artifacts. Reviewers must identify provisions that depart from policy, determine the implicated condition, and explain the basis for the decision. Although this work is essential for governance and accountability, it requires expert time and becomes difficult to scale across large document collections.

We instantiate this problem in non-disclosure agreement (NDA) review, where contracts are checked against a company's preferred positions for acceptable agreement language. Prior work has studied contract review through expert-annotated clause identification and document-level inference over contract hypotheses \citep{hendrycks2021cuad,koreeda2021contractnli}. NDAs are a practical evaluation setting because they are high-volume documents with recurring compliance issues, including broad confidentiality definitions, long survival periods, weak disclosure controls, and expansive remedies. At the same time, the core challenge is more general: applying organization-specific policy guidance to documents in an explicit, testable, and maintainable way.

LLMs can identify relevant passages, answer questions about obligations and exceptions, and compare document text against policy guidance. However, when the LLM itself acts as the reviewer, the compliance decision remains model-generated rather than governed by independently executable policy logic. This makes it difficult to ensure that extracted facts, policy conditions, and compliance labels are applied consistently across documents or model runs. Prior work has similarly highlighted both the promise and reliability limits of LLMs in legal reasoning \citep{guha2023legalbench,dahl2024largelegalfictions}. These limitations make such systems difficult to test, debug, and update without affecting the entire review pipeline.

This motivates a different use of language models for policy-grounded document review. Rather than asking an LLM to decide compliance in one step, we separate document-specific interpretation from policy application. The organization's guidance is represented as structured non-compliance rules, each specifying a document condition and the policy issue it triggers. For a new document, the model extracts rule-relevant facts from the text, while the final decision is made by checking those facts against the explicit rule layer. This makes the review logic easier to inspect, revise, and test independently of any single document.

In this paper, we present PolicyGuard, a neuro-symbolic framework for policy-grounded document compliance review (Figure~\ref{fig:overview}). PolicyGuard constructs an executable review engine from organizational policy guidance. The engine consists of formal non-compliance rules and an extraction layer that maps each rule atom to a targeted question answerable from document text. Given a target document, LLMs answer these local questions with supporting evidence, while deterministic symbolic evaluation applies the formal rules to produce the final compliance decision. This separates document interpretation from policy application, making the review process explicit, inspectable, and testable.

In summary, this paper makes the following contributions:

\begin{itemize}[itemsep=2pt, topsep=2pt, parsep=0pt, leftmargin=*]
\item We introduce a policy-to-engine construction method that converts organizational policy guidance into executable typed relational logic rules, atom-level extraction questions, and rule-level validation tests.
\item We present a neuro-symbolic inference pipeline that grounds rule atoms in document evidence using LLMs and detects non-compliance through deterministic symbolic evaluation.
\item We instantiate PolicyGuard for company-specific NDA compliance review, demonstrating its use on high-volume contract documents governed by organization-specific negotiation policies.
\end{itemize}

\section{Related Work}
\label{sec:related_work}

Much prior work on automated contract review treats the task as supervised legal language understanding over a fixed review schema. CUAD \citep{hendrycks2021cuad} identifies contract spans for predefined legal categories, while ContractNLI \citep{koreeda2021contractnli} frames NDA review as inference over recurring hypotheses. Broader legal benchmarks evaluate LLMs on legal reasoning \citep{guha2023legalbench}, and recent studies show that legal LLM outputs remain vulnerable to reliability and hallucination failures \citep{dahl2024largelegalfictions}. These works address important legal document analysis tasks, but not enterprise policy review, where the checks are defined by an organization's own evolving playbook.

More broadly, PolicyGuard relates to work on formalizing natural-language policies into executable constraints. This direction has recently been studied in the context of governing LLM-based agents, where policies are translated into symbolic or verifiable checks over agent actions. ShieldAgent \citep{chen2025shieldagent} extracts verifiable rules from safety policy documents and applies logical reasoning to enforce compliance over agent action trajectories. VeriGuard \citep{miculicich2025veriguard} similarly translates natural-language safety requirements into verified policy code that monitors agent actions before execution. These systems demonstrate the value of making policy constraints explicit and executable. However, they formalize policies primarily to constrain agent behavior. PolicyGuard instead formalizes organization-specific playbooks into auditable review engines that evaluate whether contracts comply with enterprise policy.

\begin{figure*}[t]
    \centering
    \includegraphics[width=\textwidth]{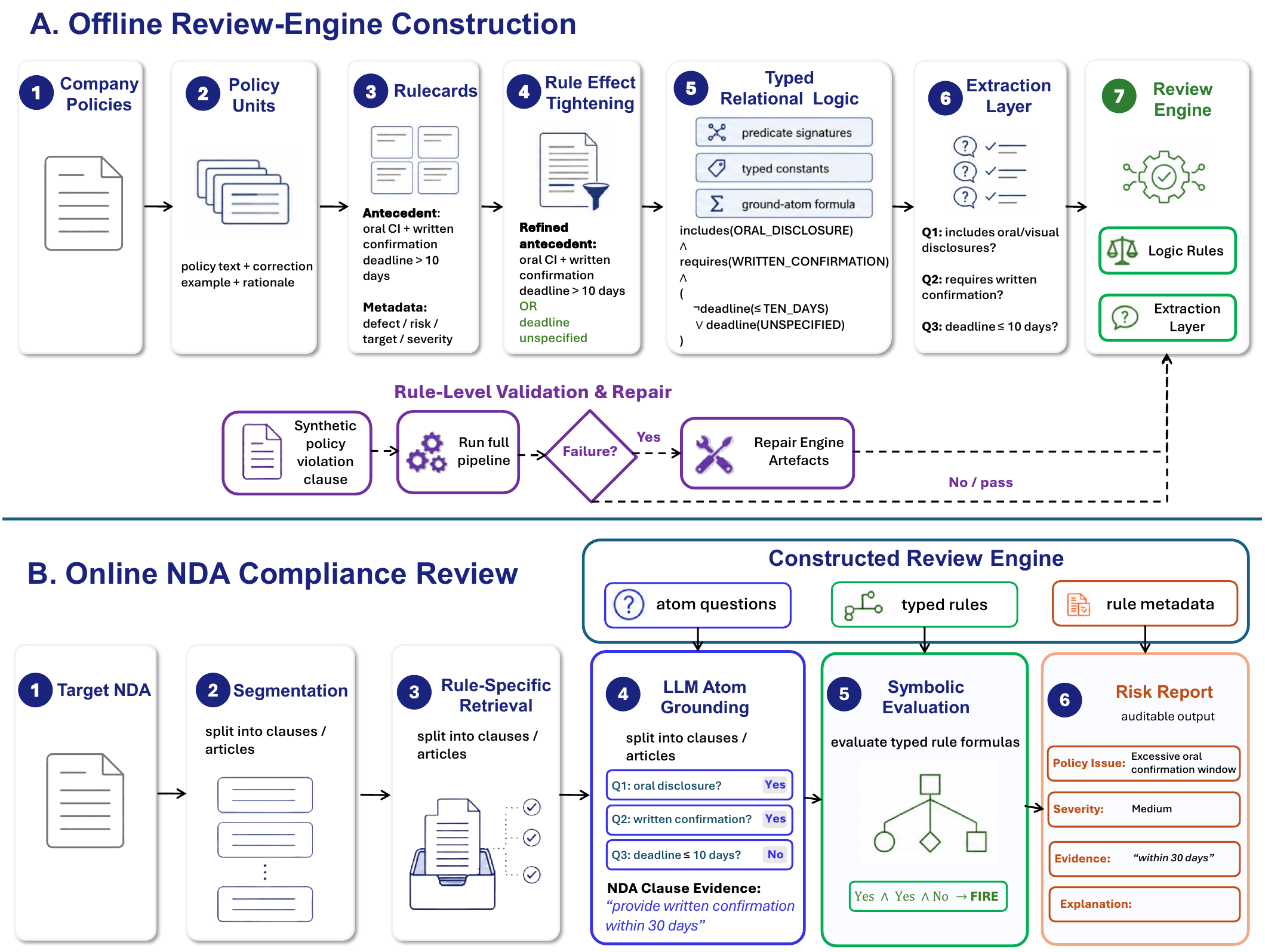}
\caption{
Overview of PolicyGuard. The figure shows the two-stage workflow: 
(A) offline construction of a policy-specific review engine from company policy guidance; and 
(B) online NDA review, where retrieved contract clauses are grounded into atom-level truth values and evaluated by the symbolic rule engine to produce an auditable risk report. 
}
\label{fig:review-engine-builder}
\end{figure*}

\section{PolicyGuard}

\subsection{Overview}
\label{sec:policyguard-overview}

PolicyGuard is a neuro-symbolic framework for policy-grounded document review. Given an organizational policy, it constructs an executable review engine that analyzes whether a target document complies with that policy. The framework is general to policy--document compliance settings; in this work, we instantiate and evaluate it on NDA review, where the target documents are contracts and the policy specifies the company's preferred compliance positions.

PolicyGuard separates policy interpretation, document interpretation, and compliance evaluation. In the build phase, policy guidance is normalized into structured rulecards and formalized as typed relational logic specifications over a shared vocabulary of canonical predicates and constants. Each rule is expressed as a Boolean formula over ground relational atoms. These atoms encode reusable policy-relevant assertions, such as actor roles, action categories, approval requirements, temporal limits, exceptions, or other domain-specific conditions. PolicyGuard also builds an extraction layer that maps each atom to a targeted true/false question answerable from the relevant document text.

The resulting review engine is validated using synthetically perturbed non-compliant examples that test whether each rule can fire through the end-to-end pipeline. When a rule fails, an LLM-based repair step revises the relevant component, such as the logic specification or extraction question. At inference time, LLMs answer local atom-level questions using retrieved document evidence, while the final compliance decision is produced by deterministic symbolic evaluation. Figure~\ref{fig:review-engine-builder} shows this end-to-end using an oral-disclosure confirmation-deadline rule.

\subsection{Building the Review Engine}
\label{sec:engine-builder}

\paragraph{Policy Formalization} This stage converts an organizational policy document into a formal policy model. PolicyGuard first decomposes the policy into self-contained policy units, each containing the policy text and contextual material needed to interpret it locally. These units are then converted into structured \emph{rulecards}. A rulecard records the policy issue, organizational position, compliance metadata, and a controlled natural-language rule specifying the document condition under which non-compliance should be detected.

Because policies often express requirements through examples, preferred positions, exceptions, or drafting patterns, initial rulecards may overfit to surface language. PolicyGuard therefore applies a rule-refinement step before formalization. This step rewrites each rulecard to capture the underlying non-compliant effect rather than only the wording used in the policy document, while preserving the same policy issue.

Each refined rulecard is then translated into a local typed relational logic specification using a structured LLM prompt. The specification contains a relational vocabulary and a rule formula. The vocabulary defines the typed predicates and constants needed to express the rule; the formula specifies the Boolean combination of ground atoms that triggers the corresponding non-compliance finding. For example, in the NDA setting, atoms may express that an obligation has an indefinite duration, that disclosure is permitted to a particular recipient category, or that prior written approval is required.

% Since rulecards are translated independently, different rules may introduce overlapping predicates or constants. PolicyGuard consolidates local vocabularies into a shared canonical vocabulary. This improves consistency across rules and allows related policy conditions to reuse the same concepts.

\paragraph{Building the Extraction Layer} The logic specifications cannot be evaluated directly on raw document text. Each ground atom must first be assigned a truth value for the relevant section, clause, or passage. PolicyGuard therefore builds an extraction layer by generating targeted true/false questions for the atoms in the logic specifications.

Each question asks whether the document text supports the relational assertion expressed by the atom. The questions are written to capture the intended policy-relevant effect rather than surface wording. At inference time, an LLM answers these questions using only the retrieved document text and provides supporting evidence for each answer. The answers are mapped to truth values for the corresponding atoms. Importantly, the LLM does not decide whether the document is compliant; it only grounds local textual assertions. The symbolic evaluator applies the formal rule to these truth values.

\paragraph{Rule-Level Validation and Repair}
PolicyGuard validates each rule with a targeted end-to-end example. For each rulecard, we create a non-compliant validation passage by rewriting a corresponding compliant reference passage while preserving its subject matter, and manually verify the resulting label. The passage is then run through atom-level extraction, truth assignment, and symbolic evaluation. If the intended non-compliance is missed, an LLM-based repair step diagnoses whether the issue lies in the logic specification or extraction layer, and revises the relevant component to better align the engine with the intended policy condition.

\subsection{Inference}
\label{sec:inference}

During inference, PolicyGuard applies the constructed review engine to a target document. The document is segmented into sections, clauses, or passages, and for each logic specification, the relevant text is retrieved. The extraction layer asks the atom-level questions for that rule and assigns truth values to the corresponding ground atoms based on the LLM's answers and supporting evidence.

The symbolic evaluator then evaluates the rule formula over these truth values. If the formula is satisfied, PolicyGuard emits the corresponding non-compliance finding together with the rule metadata and document evidence. The output includes the policy issue, severity, explanation, triggered rule, and supporting text, providing an auditable path from organizational policy to formal rule to detected non-compliance.

\section{Experiments}
\label{sec:experiments}

\begin{table*}[t]
\begin{center}
\setlength{\tabcolsep}{8pt}
\renewcommand{\arraystretch}{1.0}
\scalebox{0.82}{
\begin{tabular}{l *{10}{c} >{\columncolor{gray!10}}c >{\columncolor{gray!10}}c}
\toprule
\multirow{2}{*}{\textbf{Method}}
  & \multicolumn{2}{c}{\cellcolor{cyan!10}\textbf{NDA-1}}
  & \multicolumn{2}{c}{\cellcolor{cyan!10}\textbf{NDA-2}}
  & \multicolumn{2}{c}{\cellcolor{cyan!10}\textbf{NDA-3}}
  & \multicolumn{2}{c}{\cellcolor{cyan!10}\textbf{NDA-4}}
  & \multicolumn{2}{c}{\cellcolor{cyan!10}\textbf{NDA-5}}
  & \multicolumn{2}{c}{\cellcolor{gray!10}\textbf{Avg}} \cr
\cmidrule(lr){2-3}\cmidrule(lr){4-5}
\cmidrule(lr){6-7}\cmidrule(lr){8-9}\cmidrule(lr){10-11}
\cmidrule(lr){12-13}
  & \textbf{Acc} & \textbf{F1}
  & \textbf{Acc} & \textbf{F1}
  & \textbf{Acc} & \textbf{F1}
  & \textbf{Acc} & \textbf{F1}
  & \textbf{Acc} & \textbf{F1}
  & \textbf{Acc} & \textbf{F1} \cr
\midrule
Majority Class
  & 76.8 & -- & 86.3 & -- & 89.5 & --
  & 94.7 & -- & 85.3 & -- & 86.5 & -- \cr
\midrule
Direct Prompting
  & 76.9 & 62.1 & 73.2 & 45.7
  & 78.9 & 40.0 & 80.4 & 26.8
  & 69.6 & 36.5 & 75.8 & 42.2 \cr
Chain-of-Thought
  & 76.0 & 61.9 & 69.8 & 43.2
  & 78.4 & 42.5 & 82.4 & 29.5
  & 69.1 & 34.7 & 75.1 & 42.4 \cr
Legal Reasoning
  & 72.8 & 58.4 & 69.8 & 46.4
  & 77.5 & 46.0 & 78.8 & 24.7
  & 67.1 & 39.1 & 73.2 & 42.9 \cr
Legal Syllogism
  & 74.7 & 60.7 & 70.2 & 45.5
  & 75.3 & 40.2 & 77.7 & 26.9
  & 68.5 & 41.0 & 73.3 & 42.9 \cr
\midrule
\rowcolor{yellow!15}
\textbf{PolicyGuard}
  & \textbf{93.0} & \textbf{85.3}
  & \textbf{95.3} & \textbf{85.0}
  & \textbf{92.2} & \textbf{68.4}
  & \textbf{94.4} & \textbf{55.3}
  & \textbf{92.1} & \textbf{74.3}
  & \textbf{93.4} & \textbf{73.7} \cr
\bottomrule
\end{tabular}
}
\end{center}
\caption{\small
  NDA non-compliance clause detection performance across five contracts.
  \textbf{Acc} = Accuracy (\%), \textbf{F1} = non-compliance-class F1 (\%),
  \textbf{Avg} = macro-average.
  Majority Class predicts all clauses as compliant.
  All methods are zero-shot using GPT-4.1.
  \textbf{Bold} = best.
}
\label{tab:main_results}
\end{table*}

\begin{table}[t]
\begin{center}
\setlength{\tabcolsep}{3pt}
\renewcommand{\arraystretch}{1.0}
\scalebox{0.68}{
\begin{tabular}{l *{8}{c} >{\columncolor{gray!10}}c >{\columncolor{gray!10}}c}
\toprule
\multirow{2}{*}{\textbf{System}}
  & \multicolumn{2}{c}{\cellcolor{cyan!10}\textbf{NDA-1}}
  & \multicolumn{2}{c}{\cellcolor{cyan!10}\textbf{NDA-2}}
  & \multicolumn{2}{c}{\cellcolor{cyan!10}\textbf{NDA-3}}
  & \multicolumn{2}{c}{\cellcolor{cyan!10}\textbf{NDA-5}}
  & \multicolumn{2}{c}{\cellcolor{gray!10}\textbf{Avg}} \cr
\cmidrule(lr){2-3}\cmidrule(lr){4-5}
\cmidrule(lr){6-7}\cmidrule(lr){8-9}
\cmidrule(lr){10-11}
  & \textbf{Acc} & \textbf{F1}
  & \textbf{Acc} & \textbf{F1}
  & \textbf{Acc} & \textbf{F1}
  & \textbf{Acc} & \textbf{F1}
  & \textbf{Acc} & \textbf{F1} \cr
\midrule
\multicolumn{11}{c}{\textit{Claude Cowork w/ raw policy guidelines}} \cr
\midrule
Sonnet-4.6
  & 82.5 & 80.0 & 80.0 & 57.1 & 72.5 & 35.3 & 67.5 & 38.1 & 75.6 & 52.6 \cr
Opus-4.6
  & 82.5 & 77.4 & 92.5 & 84.2 & 85.0 & 50.0 & 80.0 & 50.0 & 85.0 & 65.4 \cr
\midrule
\multicolumn{11}{c}{\textit{PolicyGuard (Ours)}} \cr
\midrule
\rowcolor{yellow!15}
Sonnet-4.5
  & \textbf{87.5} & \textbf{83.9}
  & \textbf{87.5} & \textbf{76.2}
  & \textbf{86.8} & \textbf{61.5}
  & \textbf{86.5} & \textbf{61.5}
  & \textbf{87.1} & \textbf{70.8} \cr
\rowcolor{yellow!15}
GPT-4.1
  & \textbf{95.0} & \textbf{93.8}
  & \textbf{100.0} & \textbf{100}
  & \textbf{95.0} & \textbf{85.7}
  & \textbf{95.0} & \textbf{83.3}
  & \textbf{96.3} & \textbf{90.7} \cr
\bottomrule
\end{tabular}
}
\end{center}
\caption{\small
  Comparison against Claude Cowork on a 40-policy subset.
  \textbf{Acc} = Accuracy (\%), \textbf{F1} = non-compliance-class F1 (\%).
  PolicyGuard uses Sonnet-4.5 for model-fair comparison
  and GPT-4.1 as the primary configuration.
  NDA-4 excluded because the contract was not externally shareable.
}
\label{tab:cowork_comparison}
\end{table}

\begin{figure}[t]
  \centering
  \begin{minipage}[t]{0.49\linewidth}
    \centering
    \includegraphics[width=\linewidth]{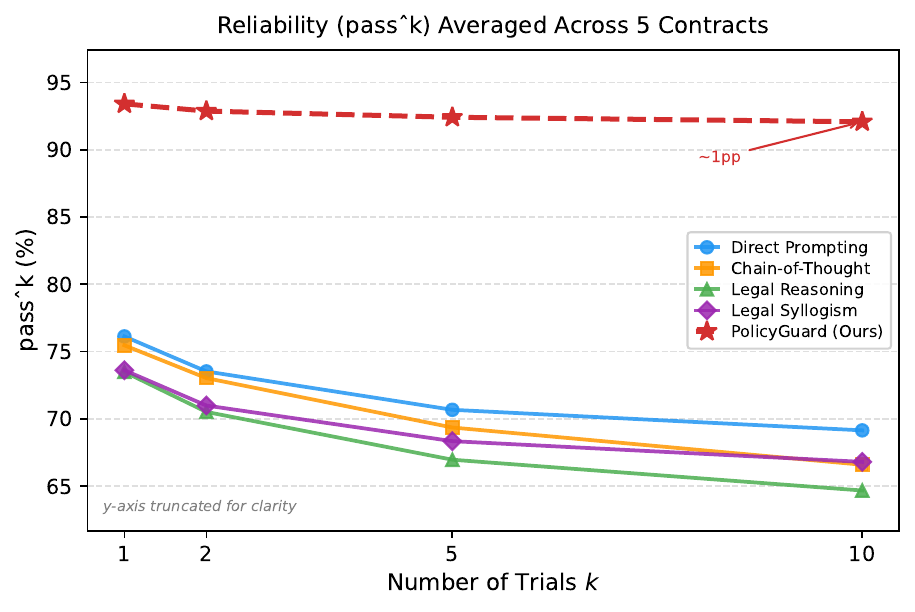}
  \end{minipage}
  \hfill
  \begin{minipage}[t]{0.49\linewidth}
    \centering
    \includegraphics[width=\linewidth]{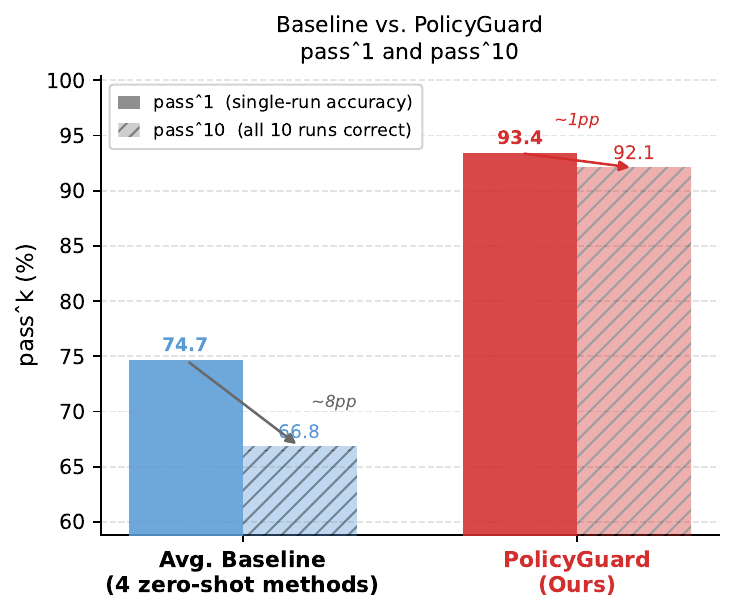}
  \end{minipage}
  \caption{Decision reliability over 10 repeated runs at temperature\,=\,0.
(\textbf{Left}) pass\^{}k curves: zero-shot baselines drop 7--9\,pp from
$k$\,=\,1 to $k$\,=\,10, while PolicyGuard drops $\sim$1\,pp.
(\textbf{Right}) pass\^{}1 vs.\ pass\^{}10 shows PolicyGuard's substantially
higher consistency for enterprise non-compliance review.}
\label{fig:reliability}
\end{figure}

\begin{table}[t]
\begin{center}
\setlength{\tabcolsep}{4pt}
\renewcommand{\arraystretch}{1.0}
\scalebox{0.75}{
\begin{tabular}{l >{\columncolor{gray!10}}c >{\columncolor{gray!10}}c >{\columncolor{gray!10}}c}
\toprule
\multirow{2}{*}{\textbf{\shortstack[l]{Backbone LLM}}}
  & \multirow{2}{*}{\textbf{Acc}}
  & \multicolumn{2}{c}{\textbf{F1}} \\
\cmidrule(lr){3-4}
  & & \textbf{Non-compliant} & \textbf{Macro} \\
\midrule
\multicolumn{4}{c}{\textit{Open-Source}} \\
\midrule
Qwen-3-8B               & 88.2 & 61.6 & 77.3 \\
Qwen-3-14B              & 90.6 & 67.2 & 80.8 \\
Qwen-3-32B              & 90.0 & 65.5 & 79.7 \\
Llama-3.3-70B           & \underline{92.3} & \underline{73.2} & \underline{84.3} \\
\midrule
\multicolumn{4}{c}{\textit{Closed-Source}} \\
\midrule
Sonnet-4.5              & 90.6 & 67.3 & 80.9 \\
GPT-4.1                 & \textbf{93.4} & \textbf{73.7} & \textbf{84.9} \\
\bottomrule
\end{tabular}
}
\end{center}
\caption{\small
  PolicyGuard generalizability across backbone LLMs,
  macro-averaged over five contracts.
  \textbf{Acc} = Accuracy (\%);
  \textbf{F1 non-compliance} = non-compliance-class F1 (\%);
  \textbf{F1 Macro} = unweighted average F1 (\%).
  Macro F1 is inflated by the ${\sim}70\%$ compliant majority;
  Non-compliance F1 is the primary metric.
  \textbf{Bold} = best overall, \underline{underline} = best open-source.
}
\label{tab:opensource_generalizability}
\end{table}

\begin{table}[t]
\begin{center}
\setlength{\tabcolsep}{4pt}
\renewcommand{\arraystretch}{1.0}
\scalebox{0.75}{
\begin{tabular}{l cc c}
\toprule
\multirow{2}{*}{\textbf{Method}}
  & \multicolumn{2}{c}{\textbf{Accuracy}}
  & \multirow{2}{*}{\textbf{Drop}} \\
\cmidrule(lr){2-3}
  & \textbf{pass\^{}1} & \textbf{pass\^{}10} & \\
\midrule
Raw Policy              & 76.1 & 69.1 & 7.0pp \\
NL Summary              & 81.4 & 76.1 & 5.3pp \\
Tightened Rule          & 82.0 & 77.2 & 4.8pp \\
Raw Policy + NL Summary & 83.0 & 77.7 & 5.3pp \\
Raw Policy + Tightened Rule & 84.4 & 79.3 & 5.1pp \\
\midrule
\rowcolor{yellow!15}
\textbf{PolicyGuard (Full)} & \textbf{93.4} & \textbf{92.1} & \textbf{1.3pp} \\
\bottomrule
\end{tabular}
}
\end{center}
\caption{\small
  Policy representation ablation using direct zero-shot prompting,
  macro-averaged over five contracts across 10 runs at temperature zero.
  \textbf{pass\^{}1} = single-run accuracy (\%);
  \textbf{pass\^{}10} = all-10-runs accuracy (\%);
  \textbf{Drop} = pass\^{}1 $-$ pass\^{}10 (pp).
  \textbf{Bold} = best.
}
\label{tab:ablation_rulecard}
\end{table}

We evaluate PolicyGuard on company-specific NDA compliance review: given a company policy guideline and a target NDA, the task is to predict whether the corresponding policy--contract pair is \textsc{Non-compliant} or \textsc{Compliant}. We use 95 guidelines from an internal playbook and five NDA contracts (NDA-3, NDA-2, NDA-1, NDA-4, NDA-5), yielding 475 policy--contract decisions verified by the company's legal team. Since the labels are imbalanced, with roughly 70\% compliant decisions, we report both accuracy and non-compliance-class F1.

\paragraph{Baselines and Models} We compare against four zero-shot prompting baselines, all receiving the full policy guideline and contract text: Direct Prompting, Chain-of-Thought \citep{kojima2022large}, Legal Reasoning (IRAC framework) \citep{yu2022legal}, and Legal Syllogism \citep{jiang-yang-2023}. We use zero-shot baselines because enterprise playbooks change frequently, making per-policy labelled examples impractical. For the main comparison (Table~\ref{tab:main_results}), all methods use GPT-4.1~\citep{openai2025gpt41}. We also evaluate PolicyGuard across six backbone LLMs (Table~\ref{tab:opensource_generalizability}), including Qwen-3~\citep{yang2025qwen3} and Llama-3.3-70B \citep{meta2024llama33}, to test portability. Full prompts are in Appendix~\ref{app:baseline_prompts}.

\paragraph{Reliability Metric} To measure deployment reliability, we adopt pass\textasciicircum{}k from \citet{yao2024taubench}, which estimates the probability that all $k$ sampled runs are correct. Even at temperature zero, repeated LLM API calls can produce different decisions due to residual non-determinism and ambiguous legal language. A production review system should return the same compliance assessment for the same policy--contract pair across runs.

\section{Results and Discussion}

\paragraph{Main Results} Table~\ref{tab:main_results} shows that PolicyGuard substantially outperforms all zero-shot baselines, improving non-compliance-class F1 by 30.8 points over the best prompting method. Although the Majority Class baseline obtains high accuracy due to label imbalance, it does not detect non-compliant cases; non-compliance-class F1 is therefore the more diagnostic metric. The prompting baselines recover some non-compliant cases but over-flag many policy--contract pairs that are compliant, yielding low precision (Appendix~\ref{app:pr_analysis}). Structured reasoning scaffolds do not meaningfully improve over Direct Prompting, suggesting that the bottleneck is not reasoning format alone, but grounding vague playbook language into explicit decision conditions. PolicyGuard flags a policy--contract pair as non-compliant only when the required predicates are grounded in contract evidence and the formal rule is satisfied.

\paragraph{Reliability} Figure~\ref{fig:reliability} reports pass\textasciicircum{}k over
ten runs at temperature zero.
The four baselines drop 6.8--8.9,pp from pass\textasciicircum{}1
to pass\textasciicircum{}10, meaning single-run accuracy
substantially overstates production reliability.
PolicyGuard drops only 1.3,pp (93.4 $\to$ 92.1), with 97.0\% of policy--contract decisions identical across all ten runs.
The gap is architectural: baselines leave the final compliance decision inside a single LLM-generated
classification step, while
PolicyGuard confines LLM variability to predicate-level
extraction and applies the final rule deterministically via
Z3~\citep{demoura2008z3}.
Per-contract breakdowns are in Appendix~\ref{app:reliability}.

\paragraph{Comparison with Claude Cowork} Table~\ref{tab:cowork_comparison} compares PolicyGuard against Claude Cowork on a 40-policy subset (NDA-4 excluded as the
contract could not be shared externally).
Cowork receives raw policy text and must infer the decision rule at review
time. On this subset, PolicyGuard with Sonnet-4.5 (87.1 Acc / 70.8 F1)
performs better than Cowork Sonnet-4.6 (75.6 / 52.6) and Cowork
Opus-4.6 (85.0 / 65.4), despite using an older backbone.
These results suggest the advantage is architectural
rather than model-driven: precompiling the playbook into an executable
engine outperforms prompting a more capable model with raw
policy text.
Cowork setup details are in
Appendix~\ref{app:cowork_appendix}.

\paragraph{Backbone Generalizability} Table~\ref{tab:opensource_generalizability} shows PolicyGuard
remains effective across open- and closed-source models.
Llama-3.3-70B (92.3 Acc / 73.2 F1) nearly matches GPT-4.1
(93.4 / 73.7), suggesting a possible path toward lower-cost or self-hosted deployment.
Notably, Qwen-3-14B outperforms the larger Qwen-3-32B in
non-compliance-class F1 (67.2 vs.\ 65.5), consistent with the
hypothesis that precise atom-level extraction rather than
open-ended reasoning capacity is the key constraint.
Accuracy and aggregate F1 can obscure performance on the minority non-compliant class;
non-compliance-class F1 is therefore the diagnostic metric throughout.

\paragraph{Ablation: Policy Representation and Symbolic
Inference} Table~\ref{tab:ablation_rulecard} isolates two sources of
gain: explicit policy representation and symbolic inference.
Making the policy more explicit progressively improves
single-run accuracy (Raw Policy 76.1 $\to$ Raw,Policy +
Tightened,Rule 84.4), validating the rulecard tightening
step.
However, no prompt-only configuration closes the reliability
gap: even the strongest variant drops 5.1,pp vs.
7.0,pp for Raw Policy, because the final compliance decision is still produced by the LLM.
The full PolicyGuard system reduces the drop to 1.3,pp by
routing the final decision through the deterministic Z3
solver; the residual variance is concentrated in a small
number of rules whose tightened conditions could be made
more precise by legal experts through the human-in-the-loop
refinement process.
Beyond accuracy, the symbolic layer provides auditability:
legal experts can inspect and edit each component
(rulecard, tightened rule, relational atoms, atom questions,
logic formula) independently, isolating errors without
affecting unrelated rules.
Retrieval-augmented baselines
(Appendix~\ref{app:retrieval_results}) show that better
context selection alone does not help: the best retrieval
F1 (42.5) matches the full-contract baseline range
(42.2--42.9), supporting the view that the bottleneck is grounding rather than
context selection alone.

\section{Conclusion}
\label{sec:conclusion}
% PolicyGuard shows that reliable enterprise policy-grounded document review requires more than prompting an LLM over policy text. By separating policy formalization, local evidence grounding, and deterministic symbolic evaluation, PolicyGuard makes document compliance review more explicit, auditable, and robust under repeated inference. In a company-specific NDA review setting with five real contracts and 95 internal policy guidelines, it substantially improves non-compliance detection and maintains near-stable decisions across repeated runs, supporting its use as a structured first-pass aid for enterprise legal workflows.

PolicyGuard shows that reliable policy-grounded document review requires more than prompting an LLM over policy text. By separating policy formalization, evidence grounding, and symbolic evaluation, PolicyGuard makes document compliance review more explicit, auditable, and robust under repeated inference. In an NDA review setting with five real contracts and 95 policy guidelines, it substantially improves non-compliance detection and maintains near-stable decisions across repeated runs. These results suggest that executable policy logic can make LLM-assisted review more controllable while preserving a link between policy, evidence, and decision.

% PolicyGuard shows that reliable enterprise policy-grounded document review requires more than prompting an LLM over policy text. By separating policy formalization, local evidence grounding, and deterministic symbolic evaluation, PolicyGuard makes document compliance review more explicit, auditable, and robust under repeated inference. In a company-specific NDA review setting with five real contracts and 95 internal policy guidelines, it substantially improves non-compliance detection and maintains near-stable decisions across repeated runs. These results suggest that executable policy logic can make LLM-assisted review more controllable while preserving a clear link between policy, evidence, and decision.

% ================================================================
%  ADD THIS as a dedicated section titled "Limitations", placed
%  after the main content and before the References, as required
%  by EMNLP 2026 Industry Track guidelines. Does not count toward
%  the page limit.
% ================================================================
\clearpage
\section*{Limitations}
\label{sec:limitations}

PolicyGuard currently represents each policy rule as a Boolean formula over atom-level facts extracted from contract text. While this makes decisions explicit and auditable, the current implementation does not construct a shared document-level fact graph linking parties, obligations, exceptions, durations, and other legal objects across the full contract, so cross-rule consistency over shared entities is not structurally enforced. The system also handles missing safeguards through targeted extraction questions rather than a first-class representation of absence or negative evidence. Our evaluation is limited to NDA risk review against one organization's internal playbook; results may not transfer directly to other organizations, policy styles, contract types, or enterprise domains without additional rule construction, validation, and expert review. Finally, because the policy guidelines and NDA contracts are proprietary, we cannot release the raw policy text, contract text, company identity, or verbatim clause excerpts. We therefore report aggregate metrics, anonymized contract identifiers, and high-level qualitative findings, which limits exact external reproducibility but reflects the practical constraints of evaluating real-world enterprise policy review systems.

\bibliographystyle{unsrtnat}
\bibliography{custom}

\clearpage

\appendix

\section{Appendix}
\label{sec:appendix}

In this appendix, we provide comprehensive technical details and methodological specifications that complement the main paper. 

In particular, this appendix contains the following:

\begin{itemize}[nosep]
    \item \hyperref[app:dataset_context]{Task Context and Dataset Provenance}
    \item \hyperref[app:reliability]{Reliability Under Repeated Inference}
    \item \hyperref[app:pr_analysis]{Precision-Recall Analysis by Contract}
    \item \hyperref[app:retrieval_results]{Retrieval-Augmented Baseline Results}
    \item \hyperref[app:worked_example]{Worked Example: Review Engine Artefacts}
    \item \hyperref[app:hitl_case_studies]{Human-in-the-Loop Rule Refinement: Case Studies}
    \item \hyperref[app:cowork_appendix]{Claude Cowork Evaluation Setup}
    \item \hyperref[app:baseline_prompts]{Baseline Benchmarking Prompts}
    \item \hyperref[app:retrieval_prompts]{Retrieval Baseline Prompts}
\end{itemize}

\section{Task Context and Dataset Provenance}
\label{app:dataset_context}

\paragraph{Task context.}
Non-disclosure agreements (NDAs) are high-volume contracts in many enterprise legal workflows. Before an NDA is approved or negotiated, reviewers must determine whether the proposed language departs from the company's preferred positions, which party bears the relevant obligation or exposure, and what type of non-compliance the deviation creates. These non-compliance may involve confidentiality scope, duration, permitted disclosures, return or destruction obligations, remedies, assignment, or other recurring NDA provisions. In practice, this review is guided by an internal playbook that records the company's preferred positions, acceptable fallback language, and known problematic clause patterns.

Reviewing an NDA against such a playbook is cognitively demanding. A reviewer must interpret the contract language, identify the company's role in the agreement, determine whether the relevant safeguard is present or absent, and decide whether the provision creates a policy-level non-compliance. This process must be repeated across many policy items and many agreements. PolicyGuard is designed for this setting: it provides a structured first-pass non-compliance triage that helps reviewers inspect policy deviations more systematically.

\paragraph{Dataset provenance.}
Our evaluation uses 95 NDA policy guidelines from an internal company playbook and five real NDA contracts from the same enterprise review setting. Each evaluation item corresponds to a policy--contract pair, yielding 475 binary decisions. Ground-truth labels were produced or verified by the company's legal and business review personnel as part of the organization's contract review process. The evaluation data is therefore not a synthetic benchmark or crowd-sourced annotation set; it reflects real policy positions and real contract-review decisions from an operating enterprise workflow.

\paragraph{Confidentiality and release constraints.}
The underlying policy playbook and NDA contracts contain confidential business and legal information. We therefore do not release the raw policy text, contract text, company identity, party names, or clause excerpts. To preserve confidentiality, the paper reports aggregate metrics, anonymized contract identifiers, and high-level qualitative findings rather than the underlying documents. This also reflects the practical constraints of evaluating real-world legal systems: enterprise policy data can be central to the contribution while still being non-releasable.

\paragraph{Intended use and human oversight.}
PolicyGuard is intended to augment human legal review, not replace it.
The system does not only output a binary non-compliance decision; it produces a
structured review report that includes the supporting contract evidence,
non-compliance severity, defect category, target actor (e.g., discloser or recipient),
non-compliance explanation, decision explanation, and suggested mitigation strategy.
This report is designed to help reviewers understand why a non-compliance was
flagged and where in the contract the relevant evidence appears.
The output is intended to be reviewed by qualified personnel before any
negotiation, approval, or legal decision is made. In this workflow, the
reviewer moves from drafting a non-compliance assessment from scratch to verifying
and correcting a structured first-pass analysis.

Human oversight is also part of the review-engine construction process.
Legal experts can inspect and revise intermediate artifacts such as
rulecards, tightened rules, atom-level questions, and formal rule
conditions. This provides a mechanism for expert control over the
system's decision logic and supports targeted correction when a policy
rule is too broad, too narrow, or insufficiently explicit.

\section{Reliability Under Repeated Inference}
\label{app:reliability}

Following \citet{yao2024taubench}, we use pass\textasciicircum{}k to measure reliability under
repeated trials. Given $n$ repeated runs for the same input and $c$ successful
runs, tau-bench estimates  pass\textasciicircum{}k as the probability that all $k$ sampled
runs are successful:
\[
  \mathrm{pass\textasciicircum{}k}
  =
  \mathbb{E}_{\mathrm{task}}
  \left[
    \frac{\binom{c}{k}}{\binom{n}{k}}
  \right].
\]
In tau-bench, stochasticity arises from language-model sampling during
multi-turn agent interactions. In our setting, the input policy and contract
are fixed and decoding temperature is set to zero; nevertheless, repeated API
calls can still yield different decisions due to residual non-determinism in
LLM inference and the model's sensitivity to ambiguous legal reasoning. We
therefore use the same estimator, but interpret pass\textasciicircum{}k as a
measure of decision reliability under repeated inference for the same
policy--contract pair.

\begin{figure}[t]
\centering
\includegraphics[width=\linewidth]{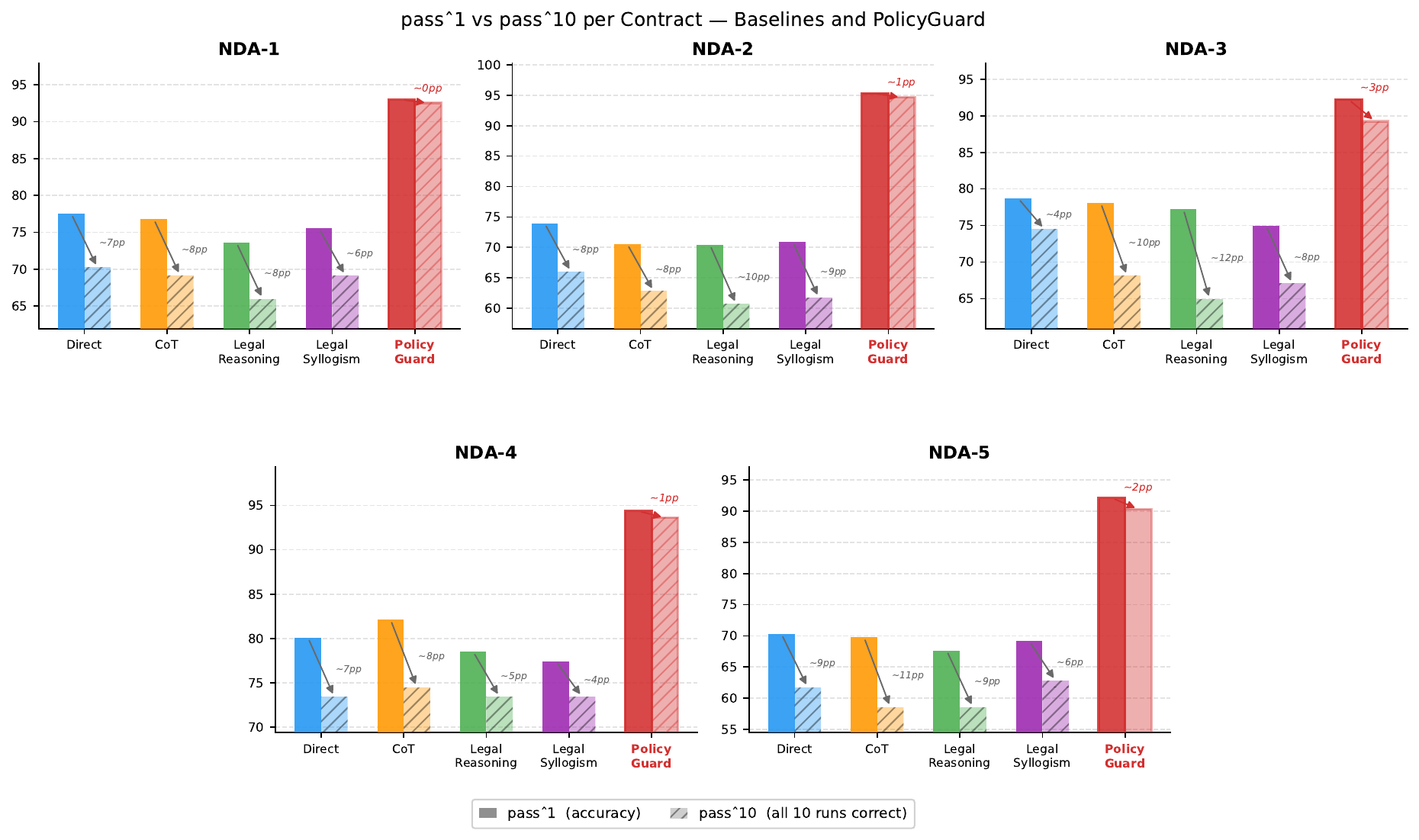}
\caption{\small
Per-contract reliability under ten repeated inference runs at temperature
zero. pass\textasciicircum{}1 measures single-run correctness, while
pass\textasciicircum{}10 requires all ten runs for a policy--contract pair
to be correct. Zero-shot baselines show larger reliability drops across
contracts, whereas PolicyGuard remains nearly stable, with the largest drop
on NDA-3.
}
\label{fig:per_contract_passk}
\end{figure}

Figure~\ref{fig:per_contract_passk} expands the aggregate reliability results
reported in the main paper by showing  pass\textasciicircum{}1 and  pass\textasciicircum{}10 separately for
each contract. The same pattern holds across all five contracts. Zero-shot
baselines show substantial reliability degradation when correctness is required
across all ten runs: Direct Prompting drops from 76.1\% to 69.1\%
($\Delta{=}7.0$\,pp), Chain-of-Thought from 75.5\% to 66.6\%
($\Delta{=}8.9$\,pp), Legal Reasoning from 73.5\% to 64.7\%
($\Delta{=}8.8$\,pp), and Legal Syllogism from 73.6\% to 66.8\%
($\Delta{=}6.8$\,pp). Thus, moderate single-run accuracy can mask a
substantial loss in repeated-run reliability.

The instability is also structured rather than uniformly distributed. Across
the zero-shot baselines, six policies are unstable in all five contracts for
at least one prompting method, indicating that repeated failures concentrate
around semantically ambiguous policy--clause pairs. Legal Reasoning shows one
of the largest drops despite its structured scaffold, reinforcing the finding
that adding explicit reasoning steps does not necessarily improve reliability
for binary compliance decisions.

PolicyGuard is substantially more stable. Its pass\textasciicircum{}1 is
93.4\% and its pass\textasciicircum{}10 is 92.1\%, a drop of only
1.3\,pp. In addition, 97.0\% of policy--contract decisions remain
identical across all ten runs. The remaining non-identical cases are localized
to a small number of ambiguous policy--clause pairs, suggesting that
PolicyGuard's residual variance is limited rather than systemic.

This reliability gap reflects the difference between direct LLM classification
and PolicyGuard's neuro-symbolic design. Zero-shot baselines require the model
to interpret legal language and emit a binary non-compliance decision in a single step,
making outputs sensitive to small variations in intermediate reasoning.
PolicyGuard separates these operations: it extracts predicate-level facts from
the contract and then applies first-order logic rules using a deterministic
solver. Since rule execution is deterministic given the extracted facts, the
remaining variance is confined to the evidence extraction stage, which operates
over narrower predicate questions rather than open-ended legal classification.
This decomposition yields more consistent decisions under repeated inference.

\section{Precision-Recall Analysis by Contract}
\label{app:pr_analysis}

Table~\ref{tab:appendix-pr} shows that all zero-shot baselines exhibit a
consistent precision--recall imbalance. Recall ranges from 55\% to 96\%
across contracts and methods, but precision remains low, typically between
15\% and 50\%, indicating systematic over-prediction of non-compliance.
This pattern is stable across all four prompting scaffolds, suggesting
that the failure mode is not specific to any particular reasoning prompt.

PolicyGuard substantially reduces this over-prediction. Precision ranges
from 47.9\% to 83.8\% across the five contracts, while recall ranges from
66.0\% to 96.2\%. The lowest precision and recall occur on NDA-4, where
the non-compliance class contains only five policies. In such a small
positive class, a few false positives or missed detections can
disproportionately affect precision, recall, and F1. NDA-1 and NDA-5 also
show recall below 90\% (86.9\% and 76.4\%, respectively), reflecting
remaining ambiguity in some policy--clause pairs. Despite this,
PolicyGuard remains competitive on recall while substantially improving
precision.

This behavior is consistent with the neuro-symbolic design. PolicyGuard
raises a non-compliance flag only when all required predicates in a
first-order logic rule are grounded in extracted contractual evidence,
preventing the speculative over-flagging observed in zero-shot prompting.

A secondary observation is that structured reasoning prompts do not
consistently outperform Direct Prompting. Additional reasoning steps may
amplify uncertainty in ambiguous policy--clause pairs rather than improve
classification, suggesting that the primary bottleneck is grounded
evidence extraction, not reasoning depth.

\begin{table}[t]
\centering
\setlength{\tabcolsep}{4pt}
\renewcommand{\arraystretch}{1.1}
\resizebox{\columnwidth}{!}{%
\begin{tabular}{l *{10}{c}}
\toprule
\multirow{2}{*}{\small\textbf{Method}}
  & \multicolumn{2}{c}{\cellcolor{cyan!10}\small\textbf{NDA-1}}
  & \multicolumn{2}{c}{\cellcolor{cyan!10}\small\textbf{NDA-2}}
  & \multicolumn{2}{c}{\cellcolor{cyan!10}\small\textbf{NDA-3}}
  & \multicolumn{2}{c}{\cellcolor{cyan!10}\small\textbf{NDA-4}}
  & \multicolumn{2}{c}{\cellcolor{cyan!10}\small\textbf{NDA-5}} \\
\cmidrule(lr){2-3}\cmidrule(lr){4-5}
\cmidrule(lr){6-7}\cmidrule(lr){8-9}\cmidrule(lr){10-11}
  & \small\textbf{P} & \small\textbf{R}
  & \small\textbf{P} & \small\textbf{R}
  & \small\textbf{P} & \small\textbf{R}
  & \small\textbf{P} & \small\textbf{R}
  & \small\textbf{P} & \small\textbf{R} \\
\midrule
\small Direct Prompting
  & \small50.2 & \small81.4
  & \small31.6 & \small82.3
  & \small29.9 & \small60.9
  & \small17.6 & \small56.7
  & \small26.4 & \small59.3 \\
\small Chain-of-Thought
  & \small49.0 & \small84.1
  & \small29.1 & \small83.8
  & \small30.6 & \small69.7
  & \small19.6 & \small60.3
  & \small25.2 & \small55.7 \\
\small Legal Reasoning
  & \small45.3 & \small82.3
  & \small30.6 & \small96.2
  & \small31.9 & \small82.7
  & \small15.9 & \small56.0
  & \small26.9 & \small71.4 \\
\small Legal Syllogism
  & \small47.5 & \small84.1
  & \small30.4 & \small90.8
  & \small28.0 & \small71.8
  & \small17.0 & \small65.0
  & \small28.3 & \small74.8 \\
\midrule
\rowcolor{yellow!15}
\small\textbf{PolicyGuard}
  & \small\textbf{83.8} & \small\textbf{86.9}
  & \small\textbf{76.2} & \small\textbf{96.2}
  & \small\textbf{60.4} & \small\textbf{79.0}
  & \small\textbf{47.9} & \small\textbf{66.0}
  & \small\textbf{72.5} & \small\textbf{76.4} \\
\bottomrule
\end{tabular}%
}
\caption{\small
  Precision (P) and Recall (R) for the non-compliant class,
  corresponding to the F1 scores reported in Table~\ref{tab:main_results}.
  All methods are zero-shot and evaluated using GPT-4.1.
  High recall with low precision across all baselines indicates
  systematic over-prediction of non-compliance.
}
\label{tab:appendix-pr}
\end{table}

\section{Retrieval-Augmented Baseline Results}
\label{app:retrieval_results}

\begin{table}[t]
\centering
\setlength{\tabcolsep}{5pt}
\renewcommand{\arraystretch}{1.1}
\resizebox{\columnwidth}{!}{%
\begin{tabular}{l cc cc}
\toprule
\multirow{2}{*}{\small\textbf{Method}}
  & \multicolumn{2}{c}{\cellcolor{cyan!10}\small\textbf{Embedding Retrieval}}
  & \multicolumn{2}{c}{\cellcolor{cyan!10}\small\textbf{LLM-guided Retrieval}} \\
\cmidrule(lr){2-3}\cmidrule(lr){4-5}
  & \small\textbf{Avg Acc} & \small\textbf{Avg F1}
  & \small\textbf{Avg Acc} & \small\textbf{Avg F1} \\
\midrule
\multicolumn{5}{c}{\cellcolor{gray!10}\textit{Top-$k$ = 1}} \\
\midrule
\small Direct Prompting    & \small 58.6 & \small 29.9 & \small 69.5 & \small 35.4 \\
\small Chain-of-Thought    & \small 54.6 & \small 28.4 & \small 67.4 & \small 37.3 \\
\small Legal Reasoning     & \small 54.5 & \small 27.5 & \small 66.9 & \small 35.8 \\
\small Legal Syllogism     & \small 59.4 & \small 31.3 & \small 69.3 & \small 38.5 \\
\midrule
\multicolumn{5}{c}{\cellcolor{gray!10}\textit{Top-$k$ = 3}} \\
\midrule
\small Direct Prompting    & \small 65.2 & \small 34.4 & \small 73.1 & \small 41.0 \\
\small Chain-of-Thought    & \small 64.8 & \small 35.7 & \small 71.5 & \small 42.5 \\
\small Legal Reasoning     & \small 61.7 & \small 32.3 & \small 70.7 & \small 40.5 \\
\small Legal Syllogism     & \small 63.8 & \small 34.8 & \small 71.4 & \small 41.3 \\
\midrule
\multicolumn{5}{c}{\cellcolor{gray!10}\textit{Top-$k$ = 5}} \\
\midrule
\small Direct Prompting    & \small 68.4 & \small 37.6 & \small 73.1 & \small 41.9 \\
\small Chain-of-Thought    & \small 67.1 & \small 36.5 & \small 72.6 & \small 42.3 \\
\small Legal Reasoning     & \small 66.3 & \small 37.0 & \small 69.6 & \small 39.4 \\
\small Legal Syllogism     & \small 66.7 & \small 38.4 & \small 72.0 & \small 42.3 \\
\bottomrule
\end{tabular}%
}
\caption{\small
  Retrieval-augmented zero-shot baseline results, macro-averaged over five
  NDA contracts. Embedding retrieval uses \texttt{all-mpnet-base-v2};
  LLM-guided retrieval uses GPT-4.1 to select the top-$k$ relevant contract
  paragraphs before applying the prompting scaffold. F1 is computed for the
  non-compliant class.
}
\label{tab:retrieval_results}
\end{table}

Table~\ref{tab:retrieval_results} reports retrieval-augmented zero-shot baseline
results using two strategies: embedding-based retrieval (\texttt{all-mpnet-base-v2})~\citep{reimers2019sbert}
and LLM-guided retrieval (GPT-4.1), evaluated at $k \in \{1,3,5\}$ across the
four prompting scaffolds and macro-averaged over the five NDA contracts.

Retrieval improves over the weakest zero-shot configurations, particularly as $k$
increases. Embedding retrieval improves from 27.5--31.3 F1 at $k{=}1$ to
36.5--38.4 F1 at $k{=}5$. LLM-guided retrieval reaches 39.4--42.5 F1 at $k{=}3$
and plateaus thereafter, suggesting that policy-aware paragraph selection largely
saturates at three retrieved chunks. Across nearly all settings, LLM-guided
retrieval outperforms embedding retrieval, indicating that selecting paragraphs by
legal intent is more effective than selection by semantic similarity.

However, retrieval does not materially exceed the full-contract zero-shot
baselines in Table~\ref{tab:main_results}. The best retrieval result (42.5 F1)
is comparable to the full-contract baseline range (42.2--42.9 F1) and far below PolicyGuard's 73.7 F1. Providing more focused context is therefore not sufficient
to resolve the over-prediction failure mode identified in
Appendix~\ref{app:pr_analysis}.

% \vspace{-0.1cm}

These results indicate that, in our setup, the bottleneck is not simply locating
relevant contract text, but converting legally ambiguous language into explicit,
verifiable conditions. Legal non-compliance depends on whether specific obligations,
exceptions, permissions, or missing safeguards are present---conditions that are
rarely stated as explicit logical propositions in NDA drafting. Retrieval exposes
the model to relevant clauses, but the model must still determine whether each
required policy predicate is satisfied. PolicyGuard addresses this grounding step
by decomposing policies into atomic facts and applying first-order logic rules
over extracted evidence, rather than relying on the model to perform this
decomposition implicitly.

\vspace{-0.2cm}

\section{Worked Example: Review Engine Artefacts}
\label{app:worked_example}

To make the construction pipeline in Section~\ref{sec:engine-builder}
concrete, we show one sanitized illustrative rule derived from the
example summarized in Figure~\ref{fig:review-engine-builder}. The
example does not reproduce verbatim policy text or contract language
from the evaluation data.

The rule concerns confidentiality definitions that include written or
oral disclosures without requiring corresponding designation safeguards.
Such language may broaden the scope of Confidential Information and
increase the Recipient's operational burden, because the parties may
later disagree about which disclosures were intended to be protected.

\paragraph{From rulecard to tightened rulecard.}
The rulecard initially expressed this condition narrowly, in terms
of the specific disclosure-type wording observed in the policy
text: written disclosure without a marking requirement, or oral
disclosure without a written-confirmation requirement.
The tightening step (Section~\ref{sec:engine-builder}) generalizes
this in two ways.
First, each surface term is broadened into a set of equivalent
phrasings the rule should recognize (e.g.,\ ``marking'' or
``legend''; ``written confirmation'' or ``written summary''),
making the rule robust to paraphrase rather than tied to one
specific wording.
Second, tightening surfaces an additional case implied by the same
underlying concern but absent from the original wording: a clause
that explicitly \emph{waives} marking requirements altogether,
rather than merely omitting them.
The resulting tightened rulecard, shown below, adds this third
disjunct while preserving the original rule's defect
classification, risk category, target party, and severity.

\begin{tcolorbox}[
    enhanced,
    breakable,
    width=\linewidth,
    title=Tightened Rulecard,
    fonttitle=\bfseries\large,
    colframe=teal!55!black,
    colback=teal!4!white,
    coltitle=white,
    colbacktitle=teal!45!black,
    boxrule=0.2mm,
    sharp corners,
    shadow={1mm}{-1mm}{0mm}{black!50!white},
    attach boxed title to top left={yshift=-3mm, xshift=3mm},
    boxed title style={sharp corners, size=small}
]

\vspace{0.3cm}

\texttt{\textbf{Rule Name:} broad\_definition\_no\_marking}

\vspace{0.2cm}
\texttt{\textbf{Goal:} Detects definitions of Confidential Information that apply to all disclosed information without requiring explicit confidentiality markings or written confirmation.}

\vspace{0.2cm}
\texttt{\textbf{Rationale:} Treating all information as confidential without specific markings imposes a additional management burden on the Recipient, creates confusion between sensitive and non-sensitive data, and increases the risk of accidental leakage.}

\vspace{0.2cm}
\texttt{━━━━━━━━━━━━━━━━━━━━━━━━━━━━}

\texttt{\textbf{ANTECEDENT (tightened):}}

\texttt{━━━━━━━━━━━━━━━━━━━━━━━━━━━━}

\texttt{IF the Definition of Confidential Information includes [Information disclosed in writing / Tangible Information] WITHOUT requiring [Confidentiality Marking / Legend]}

\texttt{OR IF the Definition of Confidential Information includes [Information disclosed orally / Visual Information / Intangible Information] WITHOUT requiring [Written Confirmation / Written Summary]}

\texttt{OR IF the Definition of Confidential Information explicitly waives [Marking Requirements / Identification Requirements]}

\vspace{0.2cm}
\texttt{━━━━━━━━━━━━━━━━━━━━━━━━━━━━}

\texttt{\textbf{CONSEQUENT:}}

\texttt{━━━━━━━━━━━━━━━━━━━━━━━━━━━━}

\texttt{Defect: OVERBROAD}\\
\texttt{Risk Category: OPERATIONAL\_BURDEN}\\
\texttt{Target Party: RECIPIENT}\\
\texttt{Severity: HIGH}

\vspace{0.2cm}
\texttt{\textbf{Explanation:} Defining Confidential Information to include unmarked or unconfirmed information expands the scope to non-sensitive or routine information data, imposing an unmanageable burden on the Recipient to protect all disclosures regardless of sensitivity.}

\end{tcolorbox}

\paragraph{From tightened rulecard to typed relational logic.}
The tightened rulecard is translated into a relational vocabulary and
a Boolean rule formula. The positive atoms identify whether the
definition includes written or oral disclosures, while the negated
atoms represent whether the corresponding designation safeguard is not
required in the retrieved clause text. In the current implementation,
these missing-safeguard conditions are evaluated through targeted
extraction questions for the local rule, rather than through a
document-level closed-world fact graph.

\begin{tcolorbox}[
    enhanced,
    breakable,
    width=\linewidth,
    title=Typed Relational Logic Specification,
    fonttitle=\bfseries\large,
    colframe=teal!55!black,
    colback=teal!4!white,
    coltitle=white,
    colbacktitle=teal!45!black,
    boxrule=0.2mm,
    sharp corners,
    shadow={1mm}{-1mm}{0mm}{black!50!white},
    attach boxed title to top left={yshift=-3mm, xshift=3mm},
    boxed title style={sharp corners, size=small}
]

\small
\vspace{0.3cm}

\texttt{\textbf{Sorts:}}

\texttt{\ \ Actor = \{DISCLOSER\}}

\texttt{\ \ Action = \{MARK, MEMORIALIZE\}}

\texttt{\ \ Object = \{WRITTEN\_DISCLOSURE,}

\texttt{\ \ \ \ \ \ \ \ \ \ ORAL\_DISCLOSURE\}}

\texttt{\ \ Container = \{CONFIDENTIALITY\_}

\texttt{\ \ \ \ \ \ \ \ \ \ \ \ \ \ DEFINITION\}}

\texttt{\ \ Constraint = \{CONFIDENTIAL\_TAG,}

\texttt{\ \ \ \ \ \ \ \ \ \ \ \ WRITTEN\_CONFIRMATION\}}

\vspace{0.2cm}
\texttt{━━━━━━━━━━━━━━━━━━━━━━━━━━━━}

\texttt{\textbf{PREDICATES:}}

\texttt{━━━━━━━━━━━━━━━━━━━━━━━━━━━━}

\texttt{\textbf{clause\_includes\_content}}

\texttt{\ \ (Container, Object) $\to$ Bool}

\texttt{\ \ Checks if a specific container (e.g.,}

\texttt{\ \ a definition) explicitly includes a}

\texttt{\ \ type of information.}

\vspace{0.15cm}
\texttt{\textbf{clause\_requires\_action}}

\texttt{\ \ (Actor, Action, Object, Constraint)}

\texttt{\ \ $\to$ Bool}

\texttt{\ \ Checks if the clause mandates a}

\texttt{\ \ specific action by an actor on an}

\texttt{\ \ object, subject to a constraint.}

\vspace{0.15cm}
\texttt{\textbf{clause\_waives\_constraint}}

\texttt{\ \ (Constraint) $\to$ Bool}

\texttt{\ \ Checks if the clause explicitly}

\texttt{\ \ waives or negates a specific}

\texttt{\ \ requirement or constraint.}

\vspace{0.2cm}
\texttt{━━━━━━━━━━━━━━━━━━━━━━━━━━━━}

\texttt{\textbf{RULE FORMULA (id: R-1):}}

\texttt{━━━━━━━━━━━━━━━━━━━━━━━━━━━━}

\texttt{(\ clause\_includes\_content(}

\texttt{\ \ \ CONFIDENTIALITY\_DEFINITION,}

\texttt{\ \ \ WRITTEN\_DISCLOSURE)}

\texttt{\ \ $\wedge$ $\neg$clause\_requires\_action(}

\texttt{\ \ \ \ DISCLOSER, MARK,}

\texttt{\ \ \ \ WRITTEN\_DISCLOSURE,}

\texttt{\ \ \ \ CONFIDENTIAL\_TAG)\ )}

\texttt{$\vee$ (\ clause\_includes\_content(}

\texttt{\ \ \ CONFIDENTIALITY\_DEFINITION,}

\texttt{\ \ \ ORAL\_DISCLOSURE)}

\texttt{\ \ $\wedge$ $\neg$clause\_requires\_action(}

\texttt{\ \ \ \ DISCLOSER, MEMORIALIZE,}

\texttt{\ \ \ \ ORAL\_DISCLOSURE,}

\texttt{\ \ \ \ WRITTEN\_CONFIRMATION)\ )}

\texttt{$\vee$ clause\_waives\_constraint(}

\texttt{\ \ \ CONFIDENTIAL\_TAG)}

\vspace{0.2cm}
\texttt{\textbf{Consequent:}}

\texttt{\ \ ProblemRaw(Defect.OVERBROAD,}

\texttt{\ \ \ Risk.OPERATIONAL\_BURDEN,}

\texttt{\ \ \ Severity.HIGH,}

\texttt{\ \ \ TargetActor.RECIPIENT)}

\end{tcolorbox}

\vspace{0.4cm}

\paragraph{From typed relational logic to extraction questions.}
Each of the five ground atoms in the formula above is mapped to a
targeted true/false question, answered by an LLM from the
retrieved clause text together with supporting evidence.

\begin{tcolorbox}[
    enhanced,
    breakable,
    width=\linewidth,
    title=Extraction-Layer Questions,
    fonttitle=\bfseries\large,
    colframe=teal!55!black,
    colback=teal!4!white,
    coltitle=white,
    colbacktitle=teal!45!black,
    boxrule=0.2mm,
    sharp corners,
    shadow={1mm}{-1mm}{0mm}{black!50!white},
    attach boxed title to top left={yshift=-3mm, xshift=3mm},
    boxed title style={sharp corners, size=small}
]

\small
\vspace{0.3cm}

\texttt{\textbf{L1} -- clause\_includes\_content(}

\texttt{\ \ CONFIDENTIALITY\_DEFINITION,}

\texttt{\ \ WRITTEN\_DISCLOSURE)}

\texttt{Review the definition of Confidential Information. Does the definition include information disclosed in writing, documents, or other tangible forms?}

\vspace{0.2cm}
\texttt{\textbf{L2} -- clause\_requires\_action(}

\texttt{\ \ DISCLOSER, MARK,}

\texttt{\ \ WRITTEN\_DISCLOSURE,}

\texttt{\ \ CONFIDENTIAL\_TAG)}

\texttt{Check the requirements for protecting written or tangible disclosures. Does the agreement require that written or tangible information be marked, stamped, or labeled with a confidentiality legend (e.g.,\ ``Confidential'') in order to qualify as Confidential Information?}

\vspace{0.2cm}
\texttt{\textbf{L3} -- clause\_includes\_content(}

\texttt{\ \ CONFIDENTIALITY\_DEFINITION,}

\texttt{\ \ ORAL\_DISCLOSURE)}

\texttt{Review the definition of Confidential Information. Does the definition include information disclosed orally, visually, or in other intangible forms?}

\vspace{0.2cm}
\texttt{\textbf{L4} -- clause\_requires\_action(}

\texttt{\ \ DISCLOSER, MEMORIALIZE,}

\texttt{\ \ ORAL\_DISCLOSURE,}

\texttt{\ \ WRITTEN\_CONFIRMATION)}

\texttt{Check the requirements for protecting oral, visual, or intangible disclosures. Does the agreement require that information disclosed orally or visually must be reduced to writing or summarized in a written notice to the receiving party in order to qualify as Confidential Information?}

\vspace{0.2cm}
\texttt{\textbf{L5} -- clause\_waives\_constraint(}

\texttt{\ \ CONFIDENTIAL\_TAG)}

\texttt{Review the scope of Confidential Information regarding marking requirements. Does the agreement waive confidentiality marking or designation requirements entirely, such that information may be treated as confidential without any marking or subsequent written designation or notification by the Discloser?}

\end{tcolorbox}

\noindent\small\textit{Each question is deliberately phrased to capture the intended legal effect of the corresponding atom (e.g.,\ ``reduced to writing or summarized in a written notice'') rather than searching for a single surface keyword. At inference time, the symbolic evaluator combines the LLM's answers to L1--L5 according to the rule formula above; the rule fires whenever a disclosure type is included in the definition without its corresponding safeguard, or whenever marking is waived outright.}

\section{Human-in-the-Loop Rule Refinement: Case Studies}
\label{app:hitl_case_studies}

A central design goal of PolicyGuard is that legal experts can inspect
and correct individual review-engine artefacts without retraining a
model or rewriting the system end-to-end. We illustrate this with three
sanitized refinement episodes from review-engine construction and
validation. The examples do not reproduce underlying policy text,
contract clauses, or rule specifications verbatim.

\paragraph{Case 1: Refining an extraction question.}
In one rule, an extraction question asked whether information had to be
marked confidential ``at the time of disclosure.'' This wording
introduced an unintended timing constraint: the model could treat a
marking requirement as absent even when the agreement required marking
by the discloser in a legally relevant way. The fix was localized to
the extraction layer: the question was edited to ask whether marking was
required by the discloser, without changing the rule formula, defect
category, or other literals. This illustrates a narrow refinement in
which the legal meaning of one atom is corrected while the surrounding
rule remains unchanged.

\paragraph{Case 2: Correcting rule-combination logic.}
In another rule, the individual literal extractions were correct, but
their Boolean combination produced a risk flag that reviewers judged
incorrect. This diagnosis separated an extraction problem from a
formula-level problem: the evidence questions were not the source of the
error, but the way their truth values were combined was too broad. The
rule formula was therefore simplified and revalidated against several
structurally distinct clause patterns, rather than only against the
single disputed example. The removed literals were not discarded from
the engine; they remained available for other rules that depend on the
same underlying facts.

\paragraph{Case 3: Human review of context-dependent risk.}
A third case involved a rule that fired correctly according to its
formal condition. However, after reviewing the broader contract context,
legal reviewers decided that the flagged clause did not require action
in that specific agreement. We therefore did not change the rule itself,
because doing so could make the rule too narrow for future contracts.
Instead, the finding was treated as a human-reviewed exception for that
contract. This illustrates that human-in-the-loop control is not limited
to editing extraction questions or formulas: reviewers can also decide
whether a triggered finding requires action in the specific contract
context.

Across these cases, the refinement process follows the same diagnostic
pattern. Before changing the review engine, the reviewer identifies
whether the problem lies in extraction, rule combination, or
case-specific legal interpretation, and then applies the narrowest
appropriate correction. This supports localized maintenance of the
review engine while preserving the audit trail from policy artefacts to
reported findings.

\section{Claude Cowork Evaluation Setup}
\label{app:cowork_appendix}

We evaluated Claude Cowork using its standard document review interface.
Because the complete 95-policy playbook could not be disclosed to an external
system, we randomly sampled 40 policies from the company playbook.
NDA-4 is excluded from this comparison because the contract was ongoing at time
of evaluation and could not be shared with external systems.

For each sampled policy, we extracted the raw Markdown guideline text,
including the policy description, compliance conditions, and non-compliance
conditions, and compiled all 40 policies into a structured configuration
file, \texttt{legal.local.md}. This file serves as the policy playbook for
Cowork and specifies the required JSON output schema, enabling responses to
be parsed and compared against ground-truth labels programmatically.
The schema matches the output fields used by PolicyGuard, so that both
systems are evaluated under the same downstream parsing and scoring protocol.
The structure of \texttt{legal.local.md} is shown below.

\vspace{0.4cm}

% ─────────────────────────────────────────────────────────────────
% BOX 1: legal.local.md
% ─────────────────────────────────────────────────────────────────
\begin{tcolorbox}[
    enhanced,
    breakable,
    width=\linewidth,
    title=\texttt{legal.local.md},
    fonttitle=\bfseries\large,
    colframe=blue!75!black,
    colback=white!10!white,
    coltitle=white,
    colbacktitle=blue!85!black,
    boxrule=0.2mm,
    sharp corners,
    shadow={1mm}{-1mm}{0mm}{black!50!white},
    attach boxed title to top left={yshift=-3mm, xshift=3mm},
    boxed title style={sharp corners, size=small}
]

\vspace{0.3cm}

\texttt{\textbf{\# Legal Policy Playbook}}

\vspace{0.15cm}

\texttt{This file defines the organization's NDA evaluation policies.}

\texttt{Each section corresponds to one policy citation. Evaluate}

\texttt{the contract clause against the Compliance and Non-compliance}

\texttt{conditions for each policy and produce the required JSON output.}

\vspace{0.1cm}

\texttt{\textbf{Total policies loaded: 40}}

\vspace{0.2cm}
\texttt{━━━━━━━━━━━━━━━━━━━━━━━━━━━━}

\texttt{\textbf{\#\# Policy [CITATION-ID]}}

\vspace{0.05cm}

\texttt{[Policy description]}

\vspace{0.15cm}

\texttt{\textbf{\#\#\# Compliance Conditions:}}

\texttt{\ \ [Raw policy compliance condition text]}

\vspace{0.1cm}

\texttt{\textbf{\#\#\# Non-compliance Conditions:}}

\texttt{\ \ [Raw policy non-compliance condition text]}

\vspace{0.15cm}

\texttt{━━━━━━━━━━━━━━━━━━━━━━━━━━━━}

\texttt{\ \ \ \ \ \ \ \ \ \ \ \ $\vdots$}

\texttt{━━━━━━━━━━━━━━━━━━━━━━━━━━━━}

\vspace{0.2cm}

\texttt{\textbf{\# Output Format Instructions}}

\vspace{0.1cm}

\texttt{For every policy citation evaluated, return ONLY the following}

\texttt{JSON structure. One object per policy citation per contract.}

\texttt{Do not add any commentary, preamble, or fields outside this}

\texttt{structure. Do not skip any policy citation.}

\vspace{0.2cm}
\texttt{━━━━━━━━━━━━━━━━━━━━━━━━━━━━}

\texttt{\{}

\texttt{"contract": "<contract filename>",}

\texttt{"policy\_citation": "<e.g. Policy [CITATION-ID]>",}

\texttt{"risk\_detected": "<yes / no>",}

\texttt{"defect": "<TOO\_STRICT / TOO\_LOOSE / NA>",}

\texttt{"severity": "<HIGH / MEDIUM / LOW / NA>",}

\texttt{"target\_actor": "<RECIPIENT / DISCLOSER / NA>",}

\texttt{"risk": "<e.g. LIABILITY\_RISK / NA>",}

\texttt{"risk\_explanation": "<one to two sentences>",}

\texttt{"nl\_decision\_explanation": "<one to two sentences>",}

\texttt{"other\_details": "<observations or null>"}

\texttt{\}}

\vspace{0.1cm}

\end{tcolorbox}

\vspace{0.5cm}

Once \texttt{legal.local.md} was placed in the shared Cowork folder
alongside the NDA contract, we issued the following evaluation instruction
directly in the Cowork interface for each contract.

% ─────────────────────────────────────────────────────────────────
% BOX 2: Cowork evaluation prompt
% ─────────────────────────────────────────────────────────────────
\begin{tcolorbox}[
    enhanced,
    breakable,
    width=\linewidth,
    title=Claude Cowork Evaluation Prompt,
    fonttitle=\bfseries\large,
    colframe=blue!75!black,
    colback=white!10!white,
    coltitle=white,
    colbacktitle=blue!85!black,
    boxrule=0.2mm,
    sharp corners,
    shadow={1mm}{-1mm}{0mm}{black!50!white},
    attach boxed title to top left={yshift=-3mm, xshift=3mm},
    boxed title style={sharp corners, size=small}
]

\vspace{0.3cm}

\texttt{Using the policy playbook and output format instructions}

\texttt{in \textbf{legal.local.md}, evaluate the NDA in}

\texttt{\textbf{[contract\_name].txt}} against all loaded policy citations.

\vspace{0.2cm}
\texttt{━━━━━━━━━━━━━━━━━━━━━━━━━━━━}

\vspace{0.1cm}

\texttt{\textbf{SAVE THE RESULTS IN THIS FOLDER:}}

\vspace{0.1cm}

\texttt{━━━━━━━━━━━━━━━━━━━━━━━━━━━━}

\vspace{0.1cm}

\texttt{\textbf{[contract\_name]\_results.md}}

\texttt{\ $\rightarrow$ One JSON block per policy citation cleanly formatted.}

\vspace{0.2cm}

\end{tcolorbox}

\vspace{0.4cm}

Cowork was instructed to read \texttt{legal.local.md}, evaluate the contract against each loaded policy citation, and write the output file to the shared folder. Each JSON block in the output file contains the fields specified in the output schema, which were parsed and compared against ground-truth labels to compute accuracy and non-compliant-class F1, as reported in Table~\ref{tab:cowork_comparison}.

During evaluation, longer policy--contract inputs occasionally reached
session limits in the Cowork interface. This practical issue suggests that
applying a general-purpose document review workspace to full-playbook NDA
review may require additional preprocessing or batching for long contracts
or large policy sets.

\section{Baseline Benchmarking Prompts}
\label{app:baseline_prompts}

The following prompts implement the four zero-shot baseline methods used for NDA non-compliant estimation. All methods receive the same policy guideline and contract text as input and must predict binary: \textsc{non-compliant} (YES $\rightarrow$ 0) or \textsc{Compliant} (NO $\rightarrow$ 1). The shared definitions and final answer instruction are embedded within each prompt.

% ─────────────────────────────────────────────────────────────────
% PROMPT 1: Direct Prompting
% ─────────────────────────────────────────────────────────────────
\begin{tcolorbox}[
    enhanced,
    breakable,
    width=\linewidth,
    title=Direct Prompting,
    fonttitle=\bfseries\large,
    colframe=blue!75!black,
    colback=white!10!white,
    coltitle=white,
    colbacktitle=blue!85!black,
    boxrule=0.2mm,
    sharp corners,
    shadow={1mm}{-1mm}{0mm}{black!50!white},
    attach boxed title to top left={yshift=-3mm, xshift=3mm},
    boxed title style={sharp corners, size=small}
]

\vspace{0.3cm}

\texttt{You are a legal expert specializing in NDA (Non-Disclosure Agreement) non-compliant assessment.}

\vspace{0.2cm}
\texttt{You are given:}

\texttt{\ \ 1. A \textbf{POLICY GUIDELINE} — defines the compliance requirement for a specific NDA clause.}

\texttt{\ \ 2. A \textbf{CONTRACT TEXT} — the full NDA document to be evaluated.}

\vspace{0.15cm}
\texttt{Your task: determine whether the contract is \textbf{NON-COMPLIANT} or \textbf{COMPLIANT} with respect to this policy.}

\vspace{0.2cm}
\texttt{\textbf{DEFINITIONS:}}
\vspace{0.15cm}

\texttt{\ \ \textbf{NON-COMPLIANT (YES)}: The contract violates, omits, or fails to satisfy the requirements defined in the policy guideline.}

\texttt{\ \ \textbf{COMPLIANT (NO)}: The contract satisfies or is consistent with the requirements defined in the policy guideline.}

\vspace{0.2cm}
\texttt{━━━━━━━━━━━━━━━━━━━━━━━━━━━━}

\texttt{\textbf{POLICY GUIDELINE:}}

\texttt{━━━━━━━━━━━━━━━━━━━━━━━━━━━━}

\texttt{[Policy guideline text]}

\vspace{0.15cm}
\texttt{━━━━━━━━━━━━━━━━━━━━━━━━━━━━}

\texttt{\textbf{CONTRACT TEXT:}}

\texttt{━━━━━━━━━━━━━━━━━━━━━━━━━━━━}

\texttt{[Full NDA contract text]}

\vspace{0.2cm}
\texttt{━━━━━━━━━━━━━━━━━━━━━━━━━━━━}

\texttt{\textbf{FINAL ANSWER INSTRUCTION:}}
\vspace{0.15cm}

\texttt{You MUST end your response with EXACTLY one of the following two lines:}

\texttt{\ \ \textbf{FINAL\_ANSWER: YES} \ $\leftarrow$ if the contract is NON-COMPLIANT with respect to this policy}

\texttt{\ \ \textbf{FINAL\_ANSWER: NO} \ \ $\leftarrow$ if the contract is COMPLIANT with respect to this policy}

\end{tcolorbox}

\vspace{0.5cm}

% ─────────────────────────────────────────────────────────────────
% PROMPT 2: Chain-of-Thought
% ─────────────────────────────────────────────────────────────────
\begin{tcolorbox}[
    enhanced,
    breakable,
    width=\linewidth,
    title=Chain-of-Thought (CoT),
    fonttitle=\bfseries\large,
    colframe=blue!75!black,
    colback=white!10!white,
    coltitle=white,
    colbacktitle=blue!85!black,
    boxrule=0.2mm,
    sharp corners,
    shadow={1mm}{-1mm}{0mm}{black!50!white},
    attach boxed title to top left={yshift=-3mm, xshift=3mm},
    boxed title style={sharp corners, size=small}
]

\vspace{0.3cm}

\texttt{You are a legal expert specializing in NDA (Non-Disclosure Agreement) non-compliant assessment.}

\vspace{0.2cm}
\texttt{You are given:}

\texttt{\ \ 1. A \textbf{POLICY GUIDELINE} — defines the compliance requirement for a specific NDA clause.}

\texttt{\ \ 2. A \textbf{CONTRACT TEXT} — the full NDA document to be evaluated.}

\vspace{0.15cm}
\texttt{Your task: determine whether the contract is \textbf{NON-COMPLIANT} or \textbf{COMPLIANT} with respect to this policy.}

\vspace{0.2cm}
\texttt{\textbf{DEFINITIONS:}}
\vspace{0.15cm}

\texttt{\ \ \textbf{NON-COMPLIANT (YES)}: The contract violates, omits, or fails to satisfy the requirements defined in the policy guideline.}

\texttt{\ \ \textbf{COMPLIANT (NO)}: The contract satisfies or is consistent with the requirements defined in the policy guideline.}

\vspace{0.2cm}
\texttt{━━━━━━━━━━━━━━━━━━━━━━━━━━━━}

\texttt{\textbf{POLICY GUIDELINE:}}

\texttt{━━━━━━━━━━━━━━━━━━━━━━━━━━━━}

\texttt{[Policy guideline text]}

\vspace{0.15cm}
\texttt{━━━━━━━━━━━━━━━━━━━━━━━━━━━━}

\texttt{\textbf{CONTRACT TEXT:}}

\texttt{━━━━━━━━━━━━━━━━━━━━━━━━━━━━}

\texttt{[Full NDA contract text]}

\vspace{0.2cm}
\texttt{━━━━━━━━━━━━━━━━━━━━━━━━━━━━}

\texttt{\textbf{INSTRUCTIONS:}}
\vspace{0.15cm}

\texttt{\ \ 1. Think through your reasoning \textbf{step by step} before reaching a conclusion.}

\texttt{\ \ 2. Identify the relevant clause(s) in the contract that bear on this policy.}

\texttt{\ \ 3. Determine whether they \textit{satisfy}, \textit{violate}, or \textit{fail to address} the policy requirement.}

\texttt{\ \ 4. You MUST end your response with EXACTLY one of the following two lines:}

\texttt{\ \ \ \ \textbf{FINAL\_ANSWER: YES} \ $\leftarrow$ if the contract is NON-COMPLIANT with respect to this policy}

\texttt{\ \ \ \ \textbf{FINAL\_ANSWER: NO} \ \ $\leftarrow$ if the contract is COMPLIANT with respect to this policy}

\vspace{0.2cm}
\texttt{Let's think step by step.}

\end{tcolorbox}

\vspace{0.5cm}

% ─────────────────────────────────────────────────────────────────
% PROMPT 3: Legal Reasoning (IRAC)
% ─────────────────────────────────────────────────────────────────
\begin{tcolorbox}[
    enhanced,
    breakable,
    width=\linewidth,
    title=Legal Reasoning (IRAC Framework),
    fonttitle=\bfseries\large,
    colframe=blue!75!black,
    colback=white!10!white,
    coltitle=white,
    colbacktitle=blue!85!black,
    boxrule=0.2mm,
    sharp corners,
    shadow={1mm}{-1mm}{0mm}{black!50!white},
    attach boxed title to top left={yshift=-3mm, xshift=3mm},
    boxed title style={sharp corners, size=small}
]

\vspace{0.3cm}

\texttt{You are a legal expert specializing in NDA (Non-Disclosure Agreement) non-compliant assessment.}

\vspace{0.2cm}
\texttt{Use the \textbf{IRAC legal reasoning framework} to determine whether the contract is \textbf{NON-COMPLIANT} or \textbf{COMPLIANT} with respect to the given policy.}

\vspace{0.2cm}
\texttt{\textbf{IRAC FRAMEWORK:}}
\vspace{0.15cm}

\texttt{\ \ \textbf{Issue} \ \ \ \ \ \ — Identify the specific compliance question raised by this policy.}

\texttt{\ \ \textbf{Rule} \ \ \ \ \ \ \ \ — State the compliance requirement as defined by the policy guideline.}

\texttt{\ \ \textbf{Application} — Apply the rule to the specific provisions found (or absent) in the contract.}

\texttt{\ \ \textbf{Conclusion} \ — State whether the contract is NON-COMPLIANT or COMPLIANT based on the above.}

\vspace{0.2cm}
\texttt{\textbf{DEFINITIONS:}}
\vspace{0.15cm}

\texttt{\ \ \textbf{NON-COMPLIANT (YES)}: The contract violates, omits, or fails to satisfy the requirements defined in the policy guideline.}

\texttt{\ \ \textbf{COMPLIANT (NO)}: The contract satisfies or is consistent with the requirements defined in the policy guideline.}

\vspace{0.2cm}
\texttt{━━━━━━━━━━━━━━━━━━━━━━━━━━━━}

\texttt{\textbf{POLICY GUIDELINE:}}

\texttt{━━━━━━━━━━━━━━━━━━━━━━━━━━━━}

\texttt{[Policy guideline text]}

\vspace{0.15cm}
\texttt{━━━━━━━━━━━━━━━━━━━━━━━━━━━━}

\texttt{\textbf{CONTRACT TEXT:}}

\texttt{━━━━━━━━━━━━━━━━━━━━━━━━━━━━}

\texttt{[Full NDA contract text]}

\vspace{0.2cm}
\texttt{━━━━━━━━━━━━━━━━━━━━━━━━━━━━}

\texttt{Apply the IRAC framework (\textit{Issue, Rule, Application, Conclusion}).}

\vspace{0.15cm}
\texttt{You MUST end your response with EXACTLY one of the following two lines:}

\texttt{\ \ \textbf{FINAL\_ANSWER: YES} \ $\leftarrow$ if the contract is NON-COMPLIANT with respect to this policy}

\texttt{\ \ \textbf{FINAL\_ANSWER: NO} \ \ $\leftarrow$ if the contract is COMPLIANT with respect to this policy}

\end{tcolorbox}

\vspace{0.5cm}

% ─────────────────────────────────────────────────────────────────
% PROMPT 4: Legal Syllogism
% ─────────────────────────────────────────────────────────────────
\begin{tcolorbox}[
    enhanced,
    breakable,
    width=\linewidth,
    title=Legal Syllogism,
    fonttitle=\bfseries\large,
    colframe=blue!75!black,
    colback=white!10!white,
    coltitle=white,
    colbacktitle=blue!85!black,
    boxrule=0.2mm,
    sharp corners,
    shadow={1mm}{-1mm}{0mm}{black!50!white},
    attach boxed title to top left={yshift=-3mm, xshift=3mm},
    boxed title style={sharp corners, size=small}
]

\vspace{0.3cm}

\texttt{You are a legal expert specializing in NDA (Non-Disclosure Agreement) non-compliant assessment.}

\vspace{0.2cm}
\texttt{Use \textbf{legal syllogistic reasoning} to determine whether the contract is \textbf{NON-COMPLIANT} or \textbf{COMPLIANT} with respect to the given policy.}

\vspace{0.2cm}
\texttt{\textbf{LEGAL SYLLOGISM STRUCTURE:}}
\vspace{0.15cm}

\texttt{\ \ \textbf{Major Premise} — The compliance rule or requirement stated in the policy guideline.}

\texttt{\ \ \textbf{Minor Premise} — The specific facts, provisions, or omissions found in the contract.}

\texttt{\ \ \textbf{Conclusion} \ \ \ — Whether the contract is NON-COMPLIANT or COMPLIANT by applying the rule to the facts.}

\vspace{0.2cm}
\texttt{\textbf{DEFINITIONS:}}
\vspace{0.15cm}

\texttt{\ \ \textbf{NON-COMPLIANT (YES)}: The contract violates, omits, or fails to satisfy the requirements defined in the policy guideline.}

\texttt{\ \ \textbf{COMPLIANT (NO)}: The contract satisfies or is consistent with the requirements defined in the policy guideline.}

\vspace{0.2cm}
\texttt{━━━━━━━━━━━━━━━━━━━━━━━━━━━━}

\texttt{\textbf{POLICY GUIDELINE:}}

\texttt{━━━━━━━━━━━━━━━━━━━━━━━━━━━━}

\texttt{[Policy guideline text]}

\vspace{0.15cm}
\texttt{━━━━━━━━━━━━━━━━━━━━━━━━━━━━}

\texttt{\textbf{CONTRACT TEXT:}}

\texttt{━━━━━━━━━━━━━━━━━━━━━━━━━━━━}

\texttt{[Full NDA contract text]}

\vspace{0.2cm}
\texttt{━━━━━━━━━━━━━━━━━━━━━━━━━━━━}

\texttt{Apply legal syllogistic reasoning (\textit{Major Premise, Minor Premise, Conclusion}).}

\vspace{0.15cm}
\texttt{You MUST end your response with EXACTLY one of the following two lines:}

\texttt{\ \ \textbf{FINAL\_ANSWER: YES} \ $\leftarrow$ if the contract is NON-COMPLIANT with respect to this policy}

\texttt{\ \ \textbf{FINAL\_ANSWER: NO} \ \ $\leftarrow$ if the contract is COMPLIANT with respect to this policy}

\end{tcolorbox}

% ═══════════════════════════════════════════════════════════════════════
% APPENDIX SECTION: Retrieval Baseline Prompts
% ═══════════════════════════════════════════════════════════════════════

\section{Retrieval Baseline Prompts}
\label{app:retrieval_prompts}

The retrieval-augmented baselines follow the same two-stage structure for both strategies: first, relevant contract paragraphs are selected; second, the retrieved chunks are passed to the reasoning model in place of the full contract text. The four reasoning prompts (Direct Prompting, Chain-of-Thought, Legal Reasoning, Legal Syllogism) are identical to those described in Appendix~\ref{app:baseline_prompts}, with one structural change: the \texttt{CONTRACT TEXT} field is replaced by a \texttt{RETRIEVED CONTRACT CLAUSES} field containing only the top-$k$ paragraphs selected by the retrieval step. The two additional prompts introduced specifically for retrieval are described below.

\vspace{0.4cm}

% ─────────────────────────────────────────────────────────────────
% PROMPT 1: Query Generation (Embedding Retrieval Path)
% ─────────────────────────────────────────────────────────────────
\begin{tcolorbox}[
    enhanced,
    breakable,
    width=\linewidth,
    title=Query Generation Prompt,
    fonttitle=\bfseries\large,
    colframe=blue!75!black,
    colback=white!10!white,
    coltitle=white,
    colbacktitle=blue!85!black,
    boxrule=0.2mm,
    sharp corners,
    shadow={1mm}{-1mm}{0mm}{black!50!white},
    attach boxed title to top left={yshift=-3mm, xshift=3mm},
    boxed title style={sharp corners, size=small}
]

\vspace{0.3cm}

\texttt{Given the following NDA policy guideline, generate a concise semantic}

\texttt{search query (2--3 sentences) that would retrieve the most relevant}

\texttt{clauses from an NDA contract needed to evaluate compliance with this policy.}

\vspace{0.2cm}

\texttt{Focus on: key legal concepts, specific obligations or rights mentioned,}

\texttt{and any conditions or thresholds defined in the policy.}

\vspace{0.2cm}
\texttt{━━━━━━━━━━━━━━━━━━━━━━━━━━━━}

\texttt{\textbf{POLICY GUIDELINE:}}

\texttt{━━━━━━━━━━━━━━━━━━━━━━━━━━━━}

\texttt{[Policy guideline text]}

\vspace{0.2cm}
\texttt{━━━━━━━━━━━━━━━━━━━━━━━━━━━━}

\texttt{\textbf{OUTPUT INSTRUCTION:}}

\texttt{━━━━━━━━━━━━━━━━━━━━━━━━━━━━}

\texttt{Output only the search query, nothing else.}

\end{tcolorbox}

\vspace{0.5cm}

\noindent\small\textit{Prior to retrieval, the contract is split into paragraphs on double-newline boundaries (minimum 50 characters per chunk) and each paragraph is encoded using \texttt{all-mpnet-base-v2}. At query time, GPT-4.1 generates a concise semantic search query from the policy guideline, which is then encoded by \texttt{all-mpnet-base-v2} for retrieval. The top-$k$ paragraphs are retrieved via cosine similarity and concatenated before being passed to the reasoning prompt (Call~2). Queries are cached per contract to avoid redundant LLM calls across the four reasoning methods.}

% ─────────────────────────────────────────────────────────────────
% PROMPT 2: LLM-Guided Retrieval (Call 1)
% ─────────────────────────────────────────────────────────────────
\begin{tcolorbox}[
    enhanced,
    breakable,
    width=\linewidth,
    title=LLM-Guided Retrieval Prompt,
    fonttitle=\bfseries\large,
    colframe=blue!75!black,
    colback=white!10!white,
    coltitle=white,
    colbacktitle=blue!85!black,
    boxrule=0.2mm,
    sharp corners,
    shadow={1mm}{-1mm}{0mm}{black!50!white},
    attach boxed title to top left={yshift=-3mm, xshift=3mm},
    boxed title style={sharp corners, size=small}
]

\vspace{0.3cm}

\texttt{You are given an NDA policy guideline and a numbered list of contract}

\texttt{paragraphs. Your task is to identify the \{topk\} most relevant}

\texttt{paragraph(s) needed to evaluate whether the contract complies with}

\texttt{this policy.}

\vspace{0.2cm}
\texttt{━━━━━━━━━━━━━━━━━━━━━━━━━━━━}

\texttt{\textbf{POLICY GUIDELINE:}}

\texttt{━━━━━━━━━━━━━━━━━━━━━━━━━━━━}

\texttt{[Policy guideline text]}

\vspace{0.15cm}
\texttt{━━━━━━━━━━━━━━━━━━━━━━━━━━━━}

\texttt{\textbf{CONTRACT PARAGRAPHS:}}

\texttt{━━━━━━━━━━━━━━━━━━━━━━━━━━━━}

\texttt{[1] [First contract paragraph]}

\texttt{[2] [Second contract paragraph]}

\texttt{\ \ \ \ $\cdots$}

\texttt{[N] [N-th contract paragraph]}

\vspace{0.2cm}
\texttt{━━━━━━━━━━━━━━━━━━━━━━━━━━━━}

\texttt{\textbf{OUTPUT FORMAT:}}

\texttt{━━━━━━━━━━━━━━━━━━━━━━━━━━━━}

\texttt{Return ONLY a JSON object in exactly this format:}

\vspace{0.1cm}

\texttt{\{\ "relevant\_chunks":\ [}

\texttt{\ \ \ \ "exact text of the most relevant paragraph",}

\texttt{\ \ \ \ "exact text of the second most relevant paragraph"}

\texttt{\ ]\}}

\vspace{0.2cm}
\texttt{━━━━━━━━━━━━━━━━━━━━━━━━━━━━}

\texttt{\textbf{RULES:}}

\texttt{━━━━━━━━━━━━━━━━━━━━━━━━━━━━}

\texttt{\ \ - Return exactly \{topk\} paragraph(s).}

\texttt{\ \ - Copy the paragraph text exactly as it appears — do not paraphrase.}

\texttt{\ \ - No explanation, no preamble, no markdown fences. JSON only.}

\end{tcolorbox}

\noindent\small\textit{The contract is split into paragraphs on double-newline boundaries (minimum 50 characters per chunk) prior to this call. The returned paragraph texts are concatenated verbatim and passed to the reasoning prompt (Call~2).}

\vspace{0.4cm}

\end{document}